\pdfoutput=1

\documentclass[twoside,british,3p]{elsarticle}
\usepackage[T1]{fontenc}
\usepackage[latin9]{inputenc}
\usepackage{xcolor}
\usepackage{pagecolor}
\usepackage{pdfcolmk}
\usepackage{units}
\usepackage{textcomp}
\usepackage{bbding}
\usepackage{amsmath}
\usepackage{amssymb}
\usepackage{graphicx}
\usepackage{setspace}
\usepackage{esvect}
\PassOptionsToPackage{normalem}{ulem}
\usepackage{ulem}
\usepackage{blindtext}
\usepackage{multicol}
\usepackage{multirow}
\usepackage{booktabs}
\usepackage{longtable}
\usepackage{tikz}
\usepackage{url}
\usepackage{hyperref}
\hypersetup{
    colorlinks=true,
    linkcolor=blue,
    filecolor=magenta,      
    urlcolor=cyan,
    pdfborderstyle={/S/U}
}

\usepackage{enumitem}
\newlist{Properties}{enumerate}{2}
\setlist[Properties]{label=Property \arabic*., font=\textbf, itemindent=*}

\makeatletter

\journal{}
\bibpunct{(}{)}{;}{a}{,}{,} 
\RequirePackage{lineno}

\usepackage{graphicx}  
\DeclareGraphicsExtensions{.pdf,.jpeg,.png,.eps}

\usepackage{dcolumn}   
\usepackage{bm}

\def\ps@pprintTitle{%
 \let\@oddhead\@empty
 \let\@evenhead\@empty
 \def\@oddfoot{}%
 \let\@evenfoot\@oddfoot}

\@ifundefined{showcaptionsetup}{}{%
 \PassOptionsToPackage{caption=false}{subfig}}
\usepackage{subfig}
\makeatother

\usepackage{babel}
\begin{document}

\begin{frontmatter}{}

\title{Shapley value-based approaches to explain the robustness of classifiers in machine learning}

\author[aff1]{Guilherme Dean Pelegrina\corref{fn1}}
\ead{guidean@unicamp.br}
\author[aff2,aff3]{Sajid Siraj}
\ead{S.Siraj@leeds.ac.uk}

\cortext[fn1]{School of Applied Sciences - University of Campinas, Limeira, Brazil}

\address[aff1]{School of Applied Sciences - University of Campinas, Limeira, Brazil}
\address[aff2]{Centre for Decision Research, Leeds University Business School, Leeds, UK}
\address[aff3]{COMSATS University Islamabad, Wah Campus, Pakistan}

\begin{abstract}
The use of algorithm-agnostic approaches is an emerging area of research for explaining the contribution of individual features towards the predicted outcome. Whilst there is a focus on explaining the prediction itself, a little has been done on explaining the robustness of these models, that is, how each feature contributes towards achieving that robustness. In this paper, we propose the use of Shapley values to explain the contribution of each feature towards the model's robustness, measured in terms of Receiver-operating Characteristics (ROC) curve and the Area under the ROC curve (AUC). With the help of an illustrative example, we demonstrate the proposed idea of explaining the ROC curve, and visualising the uncertainties in these curves. For imbalanced datasets, the use of Precision-Recall Curve (PRC) is considered more appropriate, therefore we also demonstrate how to explain the PRCs with the help of Shapley values. The explanation of robustness can help analysts in a number of ways, for example, it can help in feature selection by identifying the irrelevant features that can be removed to reduce the computational complexity. It can also help in identifying the features having critical contributions or negative contributions towards robustness.   

\end{abstract}


\end{frontmatter}{}

\section{Introduction}
\label{sec:intro}

The field of business intelligence and predictive analytics has grown by leaps and bounds in last two decades~\citep{Kumar2018,Liang2018,Zhang2021}. This growth can be attributed to a significant improvement in the performance of various prediction algorithms (in terms of their accuracy, precision, recall, etc.). However, when considering the practicality of using these prediction algorithms, the majority of them tends to be quite complex. A key contribution in this domain is the recent introduction of algorithm-agnostic explanation approaches to explain the contribution of each feature towards the overall prediction~\citep{Guidotti2018,Molnar2021,burkart2021survey,Kenny2021}. It is certainly an important achievement in predictive analytics as most of the models (specially those based on random forests~\citep{Breiman2001}, deep neural networks~\citep{LeCun2015,Goodfellow2016} and extreme gradient boosting~\citep{Chen2015,Chen2016}) were previously treated as black box models and were questioned due to their complex nature. Indeed, besides the performance, aspects such as fairness and explainability (among others) are also important when deciding which machine learning (ML) model to adopt~\citep{Kleinberg2017,Choras2020,Miller2019,Bucker2022}.

One of the widely-used algorithm-agnostic approaches to explain ML models is based on the cooperative game-theoretic concept called the Shapley value~\citep{Shapley1953}. The Shapley value approach is considered as a fair way to divide the payoff in a game among its players. In ML context, it can be used as a feature attribution method, i.e. a measure that indicates how much each feature is contributing in the ML task. Therefore, the main idea is to see the ML problem (classification, prediction, etc.) as a cooperative game such that the features cooperate in order to achieve a specific goal. This goal depends on what one would like to explain, for example,~\citet{lipovetsky2001analysis} used the Shapley values to explain the coefficient of determination in regression models. As the Shapley value calculation considers all coalitions of regressors, the obtained results were consistent even in scenarios with multicollinearity.~\citet{begley2020explainability} associated payoffs in game theory to fairness measures and applied the Shapley values to explain the impact of features when there are disparate results regarding different groups of people (women and men, blacks and whites, etc.). More recently,~\citet{Giudici2021} used the Shapley values as an explanatory approach to the Lorenz Zonoid goodness of fit. Moreover, in the famous SHAP method proposed by~\citet{lundberg2017unified}, the Shapley values were used to indicate the contribution of each feature in local predictions.

In the field of ML, improving the performance of prediction algorithms has remained the main focus for decades. The performance metrics of accuracy, precision and F-measure are considered almost mandatory when evaluating any prediction algorithm. Although there are recent proposals to estimate the contribution of each feature towards prediction, there is still a need to explain their contributions towards the robustness of these models~\citep{Wang2009,Serrano2010,Xu2014}. To assess the robustness in prediction models, the Receiver Operating Characteristic (ROC) curve and the area under the ROC curve (AUC) have been widely used in ML \citep{bradley1997use}, which has traditionally been used in the field of operational research and signal processing for long time \citep{therrien1989decision}. We contend that the Shapley-values can also be used to explain robustness of the ML models, for example, by explaining the contribution of each feature towards the AUC. Imagine a situation where we achieve a model with the AUC of 0.90. It tells us how robust the prediction model is, however, it does not explain how much each feature has contributed towards this robustness. Also, one may wish to investigate whether the contribution of each feature varies for different specificity, or does it remain consistent regardless of the specificity values.

Considering this gap, we first propose the ShapAUC method, a Shapley-based approach to explain the AUC for any ML model. For this, we assume that features join in coalitions and cooperate to achieve the AUC as a common goal. Based on the AUC calculated for all coalitions of features, we calculate the Shapley values, which indicate the marginal contribution of each feature in the model robustness. As a second contribution, namely ShapROC, we propose a way that explains the contribution of each feature towards the ROC curve. Shapley value is calculated to provide the marginal contribution of each feature at each point inside the ROC curve. As the use of Precision-Recall Curves (PRCs) is preferred for imbalanced datasets, we also propose to extend the use of Shapley values to decompose PRCs as well as to calculate the contributions towards the area under the PRC (AUPRC). Based on numerical experiments in a real dataset, we show the usefulness of our proposals in explaining the contribution of each feature towards the robustness of a model.

One of the benefits of the proposed approach is to use it in feature selection. As we provide the marginal contributions of features towards robustness, it is possible to identify insignificant features, or  more importantly, identifying features having negative contribution. By removing a feature with negative contribution, we may improve the model robustness, and by removing insignificant features, we can reduce the computational complexity of the classifier.

The rest of this paper is organised as follows. Section~\ref{sec:theory} discusses the background of our proposals, which lies in the robustness of prediction models and the use of Shapley values as a feature attribution method. In Section~\ref{sec:proposals}, we present the proposed approaches. Section~\ref{sec:example} illustrates the use of our proposals in a real dataset. Finally, in Section~\ref{sec:conclu}, we show our conclusions and future perspectives.

\section{Background} \label{sec:theory}
This section presents the theoretical background used in our proposals. Firstly, we discuss the key elements when assessing the prediction model robustness based on ROC curve and PRC. Thereafter, we discuss the use of Shapley values as a feature attribution method in ML explainability.

\subsection{Measuring the performance of classifiers} 
\begin{figure*}[h!t]
\begin{centering}
\includegraphics[width=0.60\textwidth]{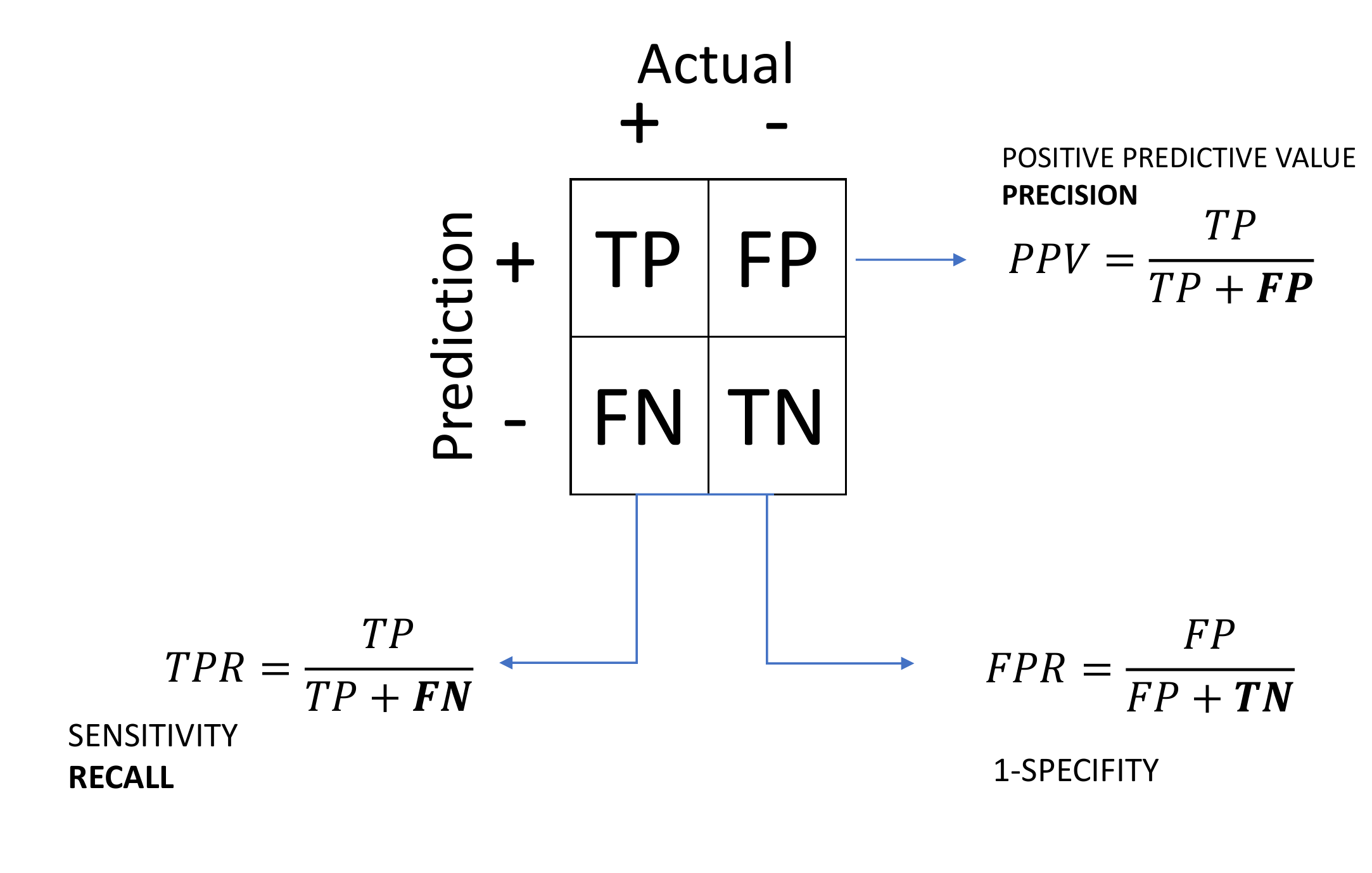} 
\par\end{centering}
\centering{}\caption{Confusion matrix showing definitions for various performance metrics.
\label{fig:confusionmatrix}}
\end{figure*}

The performance of classifiers can be measured in different ways. A common way of assessing performance is to calculate the accuracy of prediction, which is essentially a ratio of the correct predictions made out of the overall predictions carried out. In practise, this metric might not be very useful in situations where predicting positive outcomes are more important than predicting negative outcomes, or vice versa. Therefore, classification performance usually involves construction of a confusion matrix that shows the number of true positives (TPs), true negatives (TNs), false positives (FPs), and false negatives (FNs). This is illustrated in Figure \ref{fig:confusionmatrix}.

From this confusion matrix, a number of different performance measures can be obtained, for example, the ratio of true positives to actual positives is known as the true positive rate (TPR), also known as Recall or Sensitivity. Similarly, the ratio of true negatives to actual negatives is known as the true negative rate or Specificity. The complementary value of Specificity is known as false positive rate (FPR). Another way of assessing classifier is its ability to detect true positive cases out of the cases that were detected as positives. This is known as the Precision of the classifier. All these metrics capture different aspect of model's performance and are useful, however, for testing robustness of model, the construction of confusion matrix itself has to be investigated and analysed. In ML literature, it is common practise to repeat training and testing several times, and testing the consistency in these performance scores. Another common practise is to assess model's performance by varying different parameters like classification thresholds. We discuss this below in more detail.

\subsubsection{ROC curve and AUC}
\label{subsec:roc}

Receiver Operating Characteristic (ROC) curve has been widely used in machine learning \citep{bradley1997use} to assess the robustness in prediction models. In a binary classification problem, the outcome is considered positive when the prediction probability is obtained above a certain threshold. For example, in a fingerprint authentication system, a fingerprint image is scanned and a ML algorithm calculates the probability for it to be a valid fingerprint. If the predicted probability happens to be 0.40, it can be declared unauthorised considering that any value lower than 0.50 is closer to 0 (false) than 1 (true). However, we can relax this requirement by lowering the threshold value from 0.50 to 0.30, in which case, the predicted probability of 0.40 will be declared true (i.e. authorised). This means that lower threshold will have higher risk of authorising the unauthorised fingerprints (false positive), while on the other hand, raising the threshold will have higher risk of rejecting the authorised fingerprints (false negatives). Ideally, a system should be robust enough to detect all true positives regardless of the threshold values. This robustness can be investigated and quantified using ROC curve which is, simply put, a line plot between FPR and TPR values calculated by varying the threshold values for classifying predicted probability. The overall robustness of the algorithm is summarised by calculating the area under this ROC curve (AUC) which is a value between 0 and 1. A value of 1 implies that the algorithm is robust in detecting all true positives no matter what the threshold value is. 

\subsection{Precision-recall curve and AUPRC}\label{subsec:prc}

Although both ROC and AUC have been widely used, their usefulness has been debated for imbalanced classification problems~\citep{Jeni2013,Saito2015}. For example, email spam detection is a classical ML problem where the two classes are highly imbalanced~\citep{Alqatawna2015}. In this case, people prefer to minimise false positives, and therefore, only interested in the left side of the ROC curve (where FPR is close to 0), however, the calculation of AUC does not prioritise one side over other. For imbalanced classes, the use of Precision-Recall Curve (PRC) is considered more appropriate than the use of ROC and AUC~\citep{Jeni2013,Saito2015}. As the names suggests, the PRC explains the relationship between precision and recall for all threshold values. 

\subsection{Explanation in ML models}

As the use of ML is getting common, there is an increasing demand (and pressure) for the explainability of these ML models. For example, a bank's customer might ask for reasons why his/her application for credit got refused. Since the introduction of SHapley Additive exPlanations (SHAP) by \citet{lundberg2017unified}, the use of explainable AI has been introduced in many practical applications like financial risk management \citep{Bussmann2020,Bucker2022}, healthcare \citep{lundberg2018explainable,Weng2022}, inflation forecasts \citep{Aras2022}, and many more. However, so far, explainability has mostly focused on how to explain the contribution of each predictor towards attaining the predicted outcomes. While explaining the predicted outcome is an important area to investigate, it is also important to explain the robustness in predicting these values. For example, if a model is shown to have robustness of 0.85, it is important to explain how different predictors have contributed towards achieving this level of robustness. In this context, the use of Shapley values can give promising results, as it has already been demonstrated to be useful in practical applications involving ML.

\subsection{Shapley values as a feature attribution method}
\label{subsec:shapley_prop}

Before defining the Shapley value, let us first introduce the notion of cooperative games~\citep{Peleg2007}. In a cooperative game, there exists a cooperative behaviour among a set of players aiming at achieving a predetermined goal. Several practical situations can be modelled as a cooperative game problem~\citep{Wang2003,Curiel2013,Bistaffa2017}. For instance,~\citet{Kristiansen2018,He2020,Churkin2021} showed applications in power system expansion planning. In this case, different companies can cooperate in order to reduce, for instance, power losses or investment cost allocation in power transmission systems. Another example includes modelling supply chain management tasks as a cooperative game problem~\citep{Meca2003,Cachon2006,Fiestras-Janeiro2011,Zheng2019}. A common goal shared by managers can be the fixed cost paid by shipment orders. Therefore, if they form a coalition and order simultaneously, they could save more money than if they act separately.

Suppose a set $N=\left\{1, 2, \ldots, n\right\}$ of $n$ players. Mathematically, one may define a coalition game on $N$ by a characteristic function $\upsilon: \mathcal{P}(N) \rightarrow \mathbb{R}$, where $\mathcal{P}(N)$ is the power set of $N$, that maps all possible coalitions of players to real numbers, such that $\upsilon(\emptyset) = 0$. One frequently refers to $\upsilon(A)$, where $A \subseteq N$, as the payoff (or the benefit) achieved by the coalition $A$ when cooperating in the game. For example, in the supply chain task mentioned earlier, $\upsilon(A)$ could represent the savings obtained by the coalition of managers $A$ when ordering simultaneously. However, a question that arises in a cooperative game is how to divide the gains obtained by a coalition of players. One of the well-known solutions for such a sharing is called Shapley value~\citep{Shapley1953}. For each player $i \in N$, the associated Shapley value represents the marginal contribution of the player in the game payoff when considering all possible coalitions of players. It can be defined as follows:
\begin{equation}
\label{eq:shapley}
\phi_{i} = \sum_{A \subseteq N\backslash \left\{i\right\}} \frac{\left(n-\left|A\right|-1\right)!\left|A\right|!}{n!} \left[\upsilon(A \cup \left\{i\right\}) - \upsilon(A) \right],
\end{equation}
where $\left| A \right|$ indicates the cardinality of subset $A$. 

An interesting aspect of Shapley value is that it satisfies several desired properties when allocating benefits among players (see ~\citep{Young1985} for other properties and further details):

\begin{Properties}
  \item \textbf{Efficiency}: The sum of the Shapley values of all players is equal to the payoff of the grand coalition $N$ discounted by the payoff of the empty coalition. As by the definition of a game $\upsilon(\emptyset)=0$, the gain $\upsilon(N)$ is distributed among the players: \\
  \begin{equation}
  \label{eq:effic}
  \sum_{i=1}^n \phi_i = \upsilon(N) - \upsilon(\emptyset) = \upsilon(N).
  \end{equation}

  \item \textbf{Null player}:  If, for all subset $A \subseteq N$,
  \begin{equation}
  \upsilon \left(A \cup \left\{i \right\} \right) = \upsilon \left(A \right),
  \end{equation}
  then $\phi_i = 0$. 
  It means that, if there is no gain when player $i$ joins any coalition (he/she does not contribute in any payoff), he/she will not receive benefits.
  
  \item \textbf{Symmetry}: If two players $i$ and $i'$ are such that \\
  \begin{equation}
  \upsilon \left(A \cup \left\{i \right\} \right) = \upsilon \left(A \cup \left\{i' \right\} \right),
  \end{equation}
  for all $A \subset N$ which contains neither $i$ nor $i'$, then $\phi_{i} = \phi_{i'}$. 
  Therefore, if two players contribute equally when joining all coalitions, they should receive the same amount.
\end{Properties}

Given these properties, the Shapley value approach is considered to be a fair strategy to divide gains, and the use of such a solution brought attention in the explainable ML research community~\citep{lipovetsky2001analysis,lundberg2017unified,begley2020explainability,Lundberg2020,Giudici2021,Aas2021}. As mentioned in Section~\ref{sec:intro}, the main idea is to use it as a feature attribution method. Therefore, given a goal that one would like to explain (accuracy, local prediction, fairness measures, etc.), the Shapley values will indicate the contribution of each feature towards this goal. For this purpose, there are some key aspects that must be considered carefully when bringing Eq.~\eqref{eq:shapley} to the field of ML:
\begin{enumerate}
    \item Firstly, one should define $\upsilon(\cdot)$ according to what one would like to explain. For instance, if one aims at analysing the contribution of each feature towards the model's overall accuracy, one has to define $\upsilon(\cdot)$ as the accuracy of the trained model based on the TPs and FPs in the test data.
    
    \item One should be aware of what $\upsilon(\emptyset)$ represents. For example, in the shipment orders example, it is clear that there will be no savings when there is no coalition, so $\upsilon(\emptyset)$ should be zero. However, in a ML scenario, the payoff of the empty set can be a non-zero value, and therefore, calculating $\upsilon(\emptyset)$ might be more complicated in those cases. 
    
    \item Finally, another important aspect is the computational complexity of the Shapley values, as it involves retraining the model for all possible coalitions $A$, and calculating $\upsilon(A)$. This may pose a computational constraint when dealing with a high-dimensional data, as the number of payoffs exponentially increases with the number of features. In order to reduce this effort, one may consider approximation strategies that estimate the Shapley values with less computations~\citep{Strumbelj2014}.
\end{enumerate}

In the next sections, we explain how to define the aforementioned aspects and how to use the Shapley values to explain the robustness of ML classifiers. 

\section{Explaining the robustness through Shapley values}
\label{sec:proposals}

As discussed earlier, explaining the predicted outcome is an important area of research but it is also important to explain the robustness in predicting these values. Although there can be different ways to measure robustness, the use of ROC and PR curves are preferred as they span a range of threshold values to classify the predicted outcomes. This makes them independent of threshold values unlike the other measures like accuracy and F-score. Therefore, the ROC and PR curves are also considered preferred approaches for explaining the robustness of ML models. The proposed process for explaining these curves is discussed below.

\subsection{ShapAUC: Explaining the area under the ROC curve}
\label{sec:shapauc}

Assume a ROC curve obtained after training a ML model. The proposed ShapAUC method provides the contribution of each feature towards the area under this ROC curve. We can safely assume that the random classifier is the baseline for the AUC, which can be achieved even when no feature contributes towards the classifier training. In the case of a random classifier, TPRs are obviously equal to FPRs and therefore, by definition, the AUC will be 0.50. If one includes features in training, the difference between the obtained AUC and the random classifier is then explained by the contribution of such features. For example, if one achieves an AUC of 0.95, the features are contributing to improve the AUC from 0.50 (which could be obtained by a random classifier) to the actual 0.95. In other words, the marginal contribution of all features should sum up to 0.45. Based on this reasoning, we define the payoffs of ShapAUC as follows:
\begin{equation}
    \upsilon^{AUC}(A) = AUC_{A} - 0.50,
\end{equation}
where $AUC_{A}$ represents the area under the ROC curve when only the features in $A$ are used in the training step. Note that, if $A = \emptyset$ then $AUC_{\emptyset} = 0.50$ (the random classifier), and therefore $\upsilon^{AUC}(\emptyset) = AUC_{\emptyset} - 0.50 = 0$. 

Moreover, according to the efficiency property (see Eq.~\eqref{eq:effic}), $\sum_{i=1}^n \phi_i = \upsilon^{AUC}(N) - \upsilon^{AUC}(\emptyset) = AUC_{N} - 0.50$, that is, the sum of the marginal contributions of all features is equal to the difference between the AUC of the grand coalition and the AUC of the random classifier.

After retraining the ML model and calculating the payoffs for all subset of features, we interpret robustness based on the Shapley values calculated by
\begin{equation}
\label{eq:shapley_auc}
\phi_{i}^{AUC} = \sum_{A \subseteq N\backslash \left\{i\right\}} \frac{\left(n-\left|A\right|-1\right)!\left|A\right|!}{n!} \left[\upsilon^{AUC}(A \cup \left\{i\right\}) - \upsilon^{AUC}(A) \right].
\end{equation}

\begin{figure*}[h!t]
\begin{centering}
\includegraphics[width=0.81\textwidth]{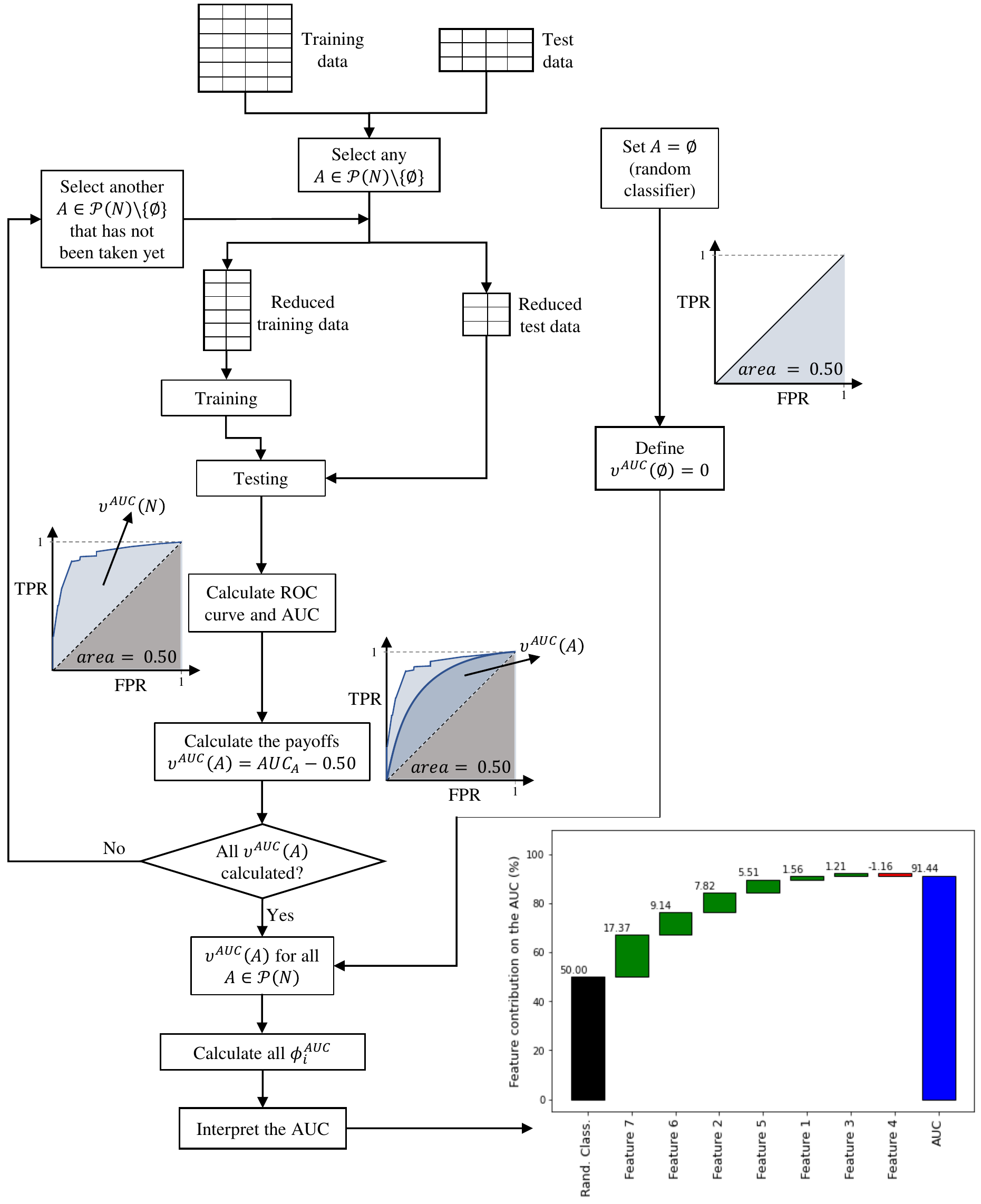} 
\par\end{centering}
\centering{}\caption{An overview of the proposed ShapAUC method to evaluate the contribution of each feature towards AUC.
\label{fig:shapauc}}
\end{figure*}

The process of ShapAUC is summarised in Figure \ref{fig:shapauc}. The procedure involves choosing a subset of features and then calculating the ROC curve for this subset, along with the AUC value. This process is then repeated for all possible subsets of features available in the ML dataset. The contribution of each feature is then estimated with the help of Eq. \ref{eq:shapley_auc}. These contributions can be visualised in a waterfall plot, as shown on the bottom right of Figure \ref{fig:shapauc}. The contribution of each feature is provided in a decreasing order (given the absolute values) that cooperates to increase the AUC from the random classifier to the actual value.

\subsection{ShapROC: Explaining the ROC curve}
\label{sec:shaproc}

In the previous subsection, we propose to explain the area under the ROC curve. It is also possible to explain the curve itself by explaining each point on the curve. The ROC curve is essentially TPR values plotted against the FPR values, so it is possible to formulate each point on the TPR curve as a coalition game. The purpose of ShapROC is to use the Shapley value as a feature attribution method to explain the TPR values in the ROC curve. We explain the process of obtaining ShapROC into three steps, as explained in the subsections below.

\subsubsection{Defining the payoffs}
\label{subsec:shaproc_payoffs}

Recall that in the case of a random classifier, the TPRs are equal to FPRs. By assuming that the random classifier is the baseline for the TPRs, the difference between a TPR and the associated FPR can be explained by the cooperation of features when they join in a coalition. For example, considering a TPR of $0.85$ for a FPR of $0.25$, the contribution of features is $0.85-0.25=0.60$. This idea leads to the following definition of payoff:
\begin{equation}
\label{eq:shaproc}
    \upsilon_{fpr}^{ROC}(A) = tpr_{fpr,A} - fpr,
\end{equation}
where $0 \leq fpr \leq 1$ and $tpr_{fpr,A}$ is the TPR associated to a given $fpr$ and a coalition $A$ of features. 

Note that, for a random classifier, $A = \emptyset$ implies that $tpr_{fpr,\emptyset} = fpr$ and, therefore, $\upsilon_{fpr}(\emptyset) = tpr_{fpr,\emptyset} - fpr = fpr - fpr = 0$. In addition, based on the efficiency property, $\sum_{i=1}^n \phi_i = \upsilon_{fpr}^{ROC}(N) - \upsilon_{fpr}^{ROC}(\emptyset) = tpr_{fpr,N} - fpr$. Therefore, by summing up the marginal contribution of each feature, we can explain the net increase in value from $fpr$ to the obtained $tpr_{fpr,N}$.

In Eq.~\eqref{eq:shaproc}, the payoffs $\upsilon_{fpr}^{ROC}(A)$ for different coalitions $A$ depend on the $fpr$ values. However, considering the fact that these values are recalculated for different classification thresholds, different coalitions might end up in generating totally different sets of FPR values, and therefore, making these coalitions incomparable to each other. To address this issue, we have to introduce an additional step for estimating TPR values in the ShapAUC method. We discuss it in the next section.

\subsubsection{Estimating the TPR values}
\label{subsec:shaproc_estimation}

We estimate the TPRs based on the standard ROC curve (calculated by using all features together). Consider that $(f_k^{ROC},t_k^{ROC})_{k=1, \cdots, l}$ represents the set of FPR and TPR values used to build the standard ROC curve. Moreover, assume that we intend to explain a specific $tpr_{fpr',N}'$ for a fixed $fpr'$ (e.g. $tpr_{0.25,N}' = 0.85$ for a fixed $fpr'=0.25$). For each $A$, as a first step, we find the nearest available FPR values on either side i.e. $f_a^{ROC},f_b^{ROC} \in \left\{ f_1^{ROC}, \ldots, f_l^{ROC} \right\}$ from $fpr'$ such that $f_a^{ROC} \leq fpr' \leq f_b^{ROC}$. 

We consider three strategies to estimate the TPR values under analysis, namely optimistic, pessimistic and interpolation strategies. The three strategies are defined below:
\begin{itemize}
    \item Optimistic strategy: $tpr_{fpr',A} = \max{(t_a^{ROC},t_b^{ROC})}$.
    \item Pessimistic strategy: $tpr_{fpr',A} = \min{(t_a^{ROC},t_b^{ROC})}$.
    \item Interpolation strategy: $tpr_{fpr',A} = \frac{(t_b^{ROC} - t_a^{ROC})(fpr' - f_a^{ROC})}{f_b^{ROC} - f_a^{ROC}} + t_a^{ROC}$.
\end{itemize}

n.b. In case of interpolation, if $f_a^{ROC} = f_b^{ROC}$, then we can simply take an average of two values as an estimate: $tpr_{fpr',A} = \frac{t_a^{ROC} + t_b^{ROC}}{2}$.

The use of optimistic strategy will provide higher values of TPRs to calculate payoffs, and therefore, it can be considered as an upper approximation of the contributions. Similarly, the pessimistic strategy will provide the lower approximation of the contributions. The interpolation strategy, on the other hand, might be considered more balanced in a way that it tends to provide a value between the upper and lower approximations. Here, the aim is not to compare these strategies or finding the most appropriate strategy, rather the aim of introducing these strategies is to demonstrate the possibility of estimating the TPR curves in order to compare results from different coalitions.

\subsubsection{Calculating the Shapley values and visualising the features contribution}
\label{subsec:shaproc_shapley}

After estimating the TPR values, the Shapley values for each slice of the curve can be calculated as follows:
\begin{equation}
\label{eq:shapley_roc}
\phi_{i}^{ROC,fpr'} = \sum_{A \subseteq N\backslash \left\{i\right\}} \frac{\left(n-\left|A\right|-1\right)!\left|A\right|!}{n!} \left[\upsilon_{fpr'}^{ROC}(A \cup \left\{i\right\}) - \upsilon_{fpr'}^{ROC}(A) \right].
\end{equation}

We can repeat the Shapley value calculations for each FPR in the ROC plot, and therefore, generating a set of curves representing contribution of each feature throughout the curve. This idea can be quite useful for ML analysts to assess the impact of each feature for different FPR values.  

\begin{figure*}[h!t]
\begin{centering}
\includegraphics[width=0.81\textwidth]{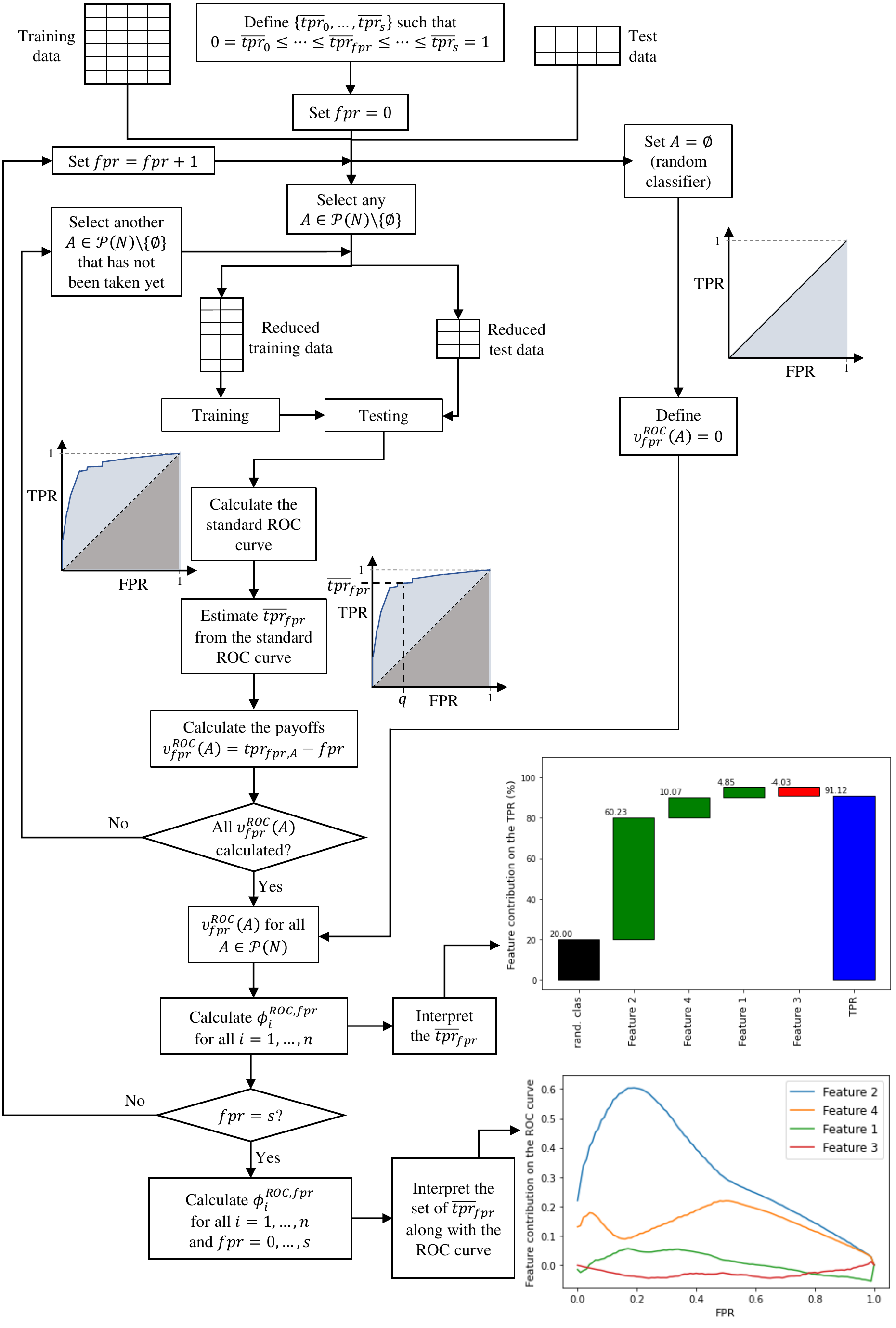} 
\par\end{centering}
\centering{}\caption{An overview of the proposed ShapROC method to evaluate the contribution of each feature towards TPR values in the ROC curve.
\label{fig:shaproc}}
\end{figure*}

Figure~\ref{fig:shaproc} illustrates the proposed ShapROC method and the features contribution visualisation towards the TPRs. In the waterfall plot, we can see how features contribute to increase the TPR from the random classifier (e.g. TPR when $fpr = 0.20$) to the actual value (e.g. $\bar{tpr}_{fpr} = 0.91$). In the figure at the bottom right, the contributions of each feature are visible for the whole range of FPR values (varying from 0 to 1).

\subsection{Relation between ShapAUC and ShapROC}
\label{sec:shaproc_shapauc}

The AUC can be approximated by the Riemann integral, that is, by the sum of very small rectangular areas calculated from the TPR and FPR values in the ROC curve. Consider a set of TPR and FPR values $(f_k^{ROC},t_k^{ROC})_{k=1, \cdots, l}$ such that $0 = f_0^{ROC} < \ldots < f_{k-1}^{ROC} < f_k^{ROC} < \ldots < f_l^{ROC} = 1$ and the difference between any $f_k^{ROC}$ and $f_{k-1}^{ROC}$ is very small. Based on the proposed ShapROC and in the efficiency property, we have that
\begin{equation}
\label{eq:relation1}
    \begin{split}
        AUC & \approx \sum_{k=2}^l \frac{\left(t_k^{ROC} + t_{k-1}^{ROC} \right)\left(f_k^{ROC} - f_{k-1}^{ROC} \right)}{2} \\
        & = \sum_{k=2}^l \frac{\left(t_k^{ROC} + t_{k-1}^{ROC} \right)\left(f_k^{ROC} - f_{k-1}^{ROC} \right)}{2} - \left( \sum_{k=2}^l \frac{\left(f_k^{ROC} + f_{k-1}^{ROC} \right)\left(f_k^{ROC} - f_{k-1}^{ROC} \right)}{2} - 0.50 \right) \\
        & = 0.50 + \sum_{k=2}^l \frac{\left(t_k^{ROC} - f_k^{ROC} + t_{k-1}^{ROC} - f_{k-1}^{ROC} \right)\left(f_k^{ROC} - f_{k-1}^{ROC} \right)}{2} \\
        & = 0.50 + \sum_{k=2}^l \frac{\left(\left(\sum_{i=1}^n \phi_{i}^{ROC,f_k^{ROC}} \right) + \left(\sum_{i=1}^n \phi_{i}^{ROC,f_{k-1}^{ROC}} \right) \right)\left(f_k^{ROC} - f_{k-1}^{ROC} \right)}{2} \\
        & = 0.50 + \sum_{k=2}^l \frac{\left(\left(\sum_{i=1}^n \phi_{i}^{ROC,f_k^{ROC}} \left(f_k^{ROC} - f_{k-1}^{ROC} \right) \right) + \left(\sum_{i=1}^n \phi_{i}^{ROC,f_{k-1}^{ROC}} \left(f_k^{ROC} - f_{k-1}^{ROC} \right) \right) \right)}{2} \\
        & = 0.50 + \sum_{i=1}^n \frac{\left(\left(\sum_{k=2}^l \phi_{i}^{ROC,f_k^{ROC}} \left(f_k^{ROC} - f_{k-1}^{ROC} \right) \right) + \left(\sum_{k=2}^l \phi_{i}^{ROC,f_{k-1}^{ROC}} \left(f_k^{ROC} - f_{k-1}^{ROC} \right) \right) \right)}{2} \\
        & = 0.50 + \sum_{i=1}^n \frac{\left(\sum_{k=2}^l \left( \phi_{i}^{ROC,f_k^{ROC}} \left(f_k^{ROC} - f_{k-1}^{ROC} \right) + \phi_{i}^{ROC,f_{k-1}^{ROC}} \left(f_k^{ROC} - f_{k-1}^{ROC} \right) \right) \right)}{2} \\
        & = 0.50 + \sum_{i=1}^n \sum_{k=2}^l \frac{ \left( \phi_{i}^{ROC,f_k^{ROC}} - \phi_{i}^{ROC,f_{k-1}^{ROC}} \right) \left(f_k^{ROC} - f_{k-1}^{ROC} \right)}{2}.
    \end{split}
\end{equation}

As we proposed in Section~\ref{sec:shapauc}, the achieved AUC can be represented as the sum of the random classifier (50\%) and the marginal contributions of features.  Mathematically, we have that
\begin{equation}
\label{eq:relation2}
    AUC = 0.50 + \sum_{i=1}^n \phi_{i}^{AUC}.
\end{equation}
In other words, we can say that we may decompose the AUC by the random classifier and the Shapley values $\phi_{i}^{AUC}$, $i=1, \ldots, n$. Therefore, each $\phi_{i}^{AUC}$ represents a ``piece of area'' from the AUC. By making a parallel between Eq.~\eqref{eq:relation1} and Eq.~\eqref{eq:relation2}, one achieves
\begin{equation}
    AUC = 0.50 + \sum_{i=1}^n \phi_{i}^{AUC} \approx 0.50 + \sum_{i=1}^n \sum_{k=2}^l \frac{ \left( \phi_{i}^{ROC,f_k^{ROC}} - \phi_{i}^{ROC,f_{k-1}^{ROC}} \right) \left(f_k^{ROC} - f_{k-1}^{ROC} \right)}{2}.
\end{equation}
Therefore, the relation between the proposed ShapAUC and ShapROC approaches are given by the following equation:
\begin{equation}
\label{eq:relation3}
    \phi_{i}^{AUC} \approx \sum_{k=2}^l \frac{ \left( \phi_{i}^{ROC,f_k^{ROC}} - \phi_{i}^{ROC,f_{k-1}^{ROC}} \right) \left(f_k^{ROC} - f_{k-1}^{ROC} \right)}{2}.
\end{equation}
Note that the approximation in Eq.~\eqref{eq:relation3} is also a sum of small areas. Indeed, as can be visualised in Figure~\ref{fig:shaproc} when interpreting the TPR values along with the ROC curve, the area under the contribution of each feature $i$ is an approximation for its contribution in the AUC. If we sum all these areas, we achieve an approximation for the AUC.

\subsection{ShapPRC and ShapAUPRC: Explaining the Precision-recall curve and the area under this curve}
\label{sec:shapprc}

As highlighted in Section~\ref{subsec:prc}, the use of Precision-Recall Curve is considered more appropriate than the use of ROC and AUC in scenarios with highly imbalanced datasets. We can also extended the same idea to use Shapley values for explaining the PRC and the AUPRC, termed as ShapPRC and ShapAUPRC, respectively. The proposed ShapAUPRC provides an explanation for the area under the PRC. When assuming a random classifier, the obtained Precision is equal to 0.50 regardless of the Recall values. Therefore, similarly as in the ShapAUC, we have the baseline area of 0.50. We then explain the contributions of features that can potentially improve the AUPRC (from the random classifier). In this case, the payoffs can be defined as follows:
\begin{equation}
    \upsilon^{AUPRC}(A) = AUPRC_{A} - 0.50,
\end{equation}
where $AUPRC_{A}$ represents the area under the PRC when only the features in $A$ are used in the training step.

The Shapley values for AUPRC can also be estimated using the steps defined for the ShapAUC. The equation for calculating the Shapley values for AUPRC can be defined as below:

\begin{equation}
\label{eq:shapley_auprc}
\phi_{i}^{AUPRC} = \sum_{A \subseteq N\backslash \left\{i\right\}} \frac{\left(n-\left|A\right|-1\right)!\left|A\right|!}{n!} \left[\upsilon^{AUPRC}(A \cup \left\{i\right\}) - \upsilon^{AUPRC}(A) \right].
\end{equation}

In ShapPRC, we propose to explain the contributions of features towards the Precision values along with the PRC. If one takes a single slice in the PRC, we can also use the ShapPRC to explain the improvement in the Precision value (from the random classifier). The steps are as defined for the ShapROC, with a redefinition of the baseline, payoffs and Shapley values calculation. For all Precision values in the PRC, the baseline remains to be $0.50$ regardless of the Recall values. This leads us to the following definition of payoffs:
\begin{equation}
\label{eq:shapprc}
    \upsilon_{rec}^{PRC}(A) = pre_{rec,A} - 0.50,
\end{equation}
where $0 \leq rec \leq 1$ and $pre_{rec,A}$ is the Precision value associated with a given Recall value ($rec$) and a coalition $A$ of features. 

When analysing the Precision values along with the PRC, we provide an explanation for each slice in it. So the equation for calculating the Shapley values for each Precision value $pre_{rec',A}'$ (associated with a Recall value $rec'$) can be defined as:
\begin{equation}
\label{eq:shapley_prc}
\phi_{i}^{PRC,rec'} = \sum_{A \subseteq N\backslash \left\{i\right\}} \frac{\left(n-\left|A\right|-1\right)!\left|A\right|!}{n!} \left[\upsilon_{rec'}^{PRC}(A \cup \left\{i\right\}) - \upsilon_{rec'}^{PRC}(A) \right].
\end{equation}

\section{Illustrative example for explaining robustness}
\label{sec:example}

We illustrate the proposed approaches\footnote{All codes and dataset can be accessed in \url{https://github.com/shaprob/shaproc}.} based on a real dataset called Banknote Authentication Dataset \citep{Lohweg2009notes}. This dataset consists of 1372 images that were used to evaluate an authentication procedure for genuine (and forged) banknotes. Wavelet Transforms were applied to these images in order to extract the following (continuous) attributes: \emph{variance}, \emph{skewness}, \emph{kurtosis} and \emph{entropy}. 

\begin{figure}[!htbp]
\centering
\subfloat[Standard ROC curve.]{\includegraphics[width=3.3in]{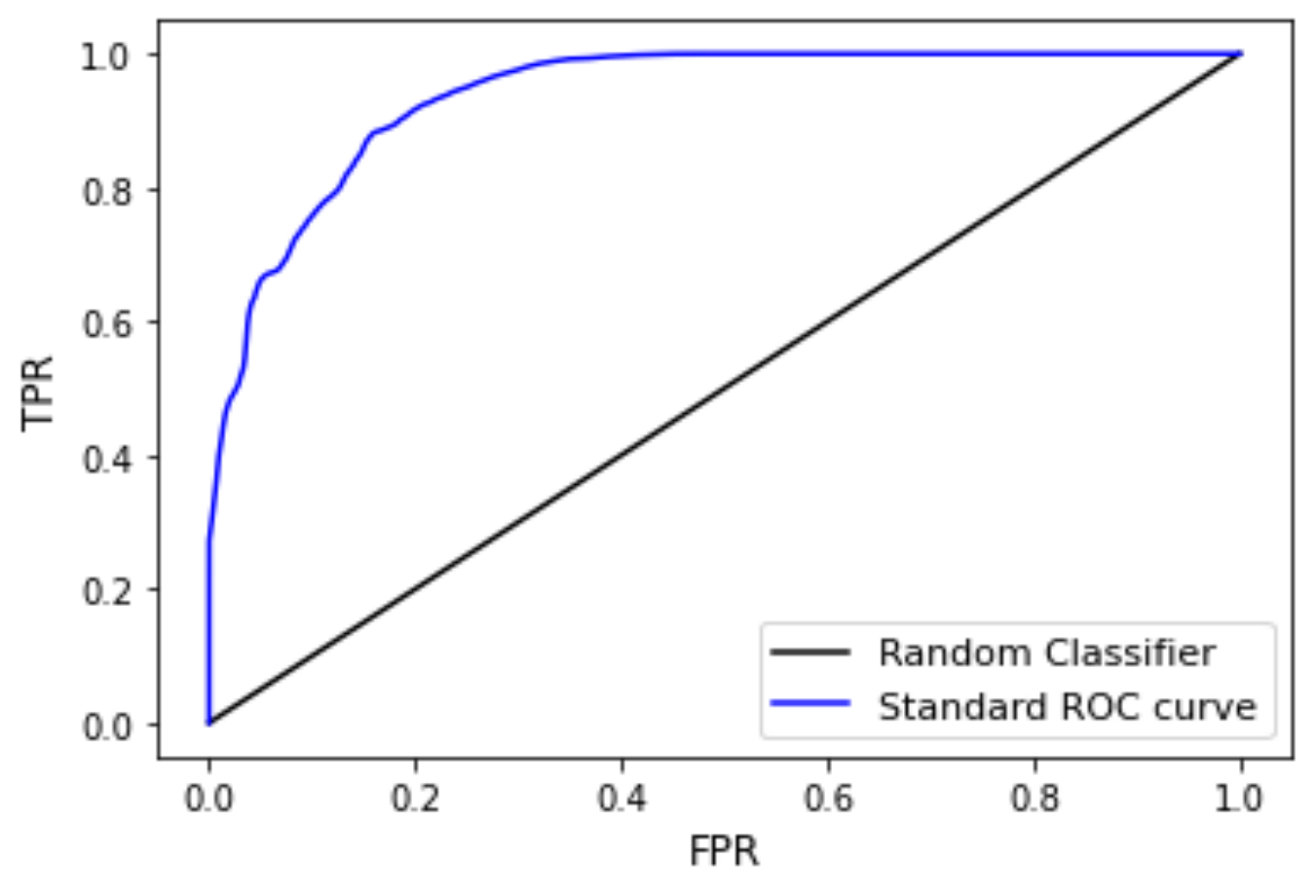}
\label{fig:robustness_stand_roc}}
\hfill
\subfloat[Contributions towards the AUC.]{\includegraphics[width=3.0in]{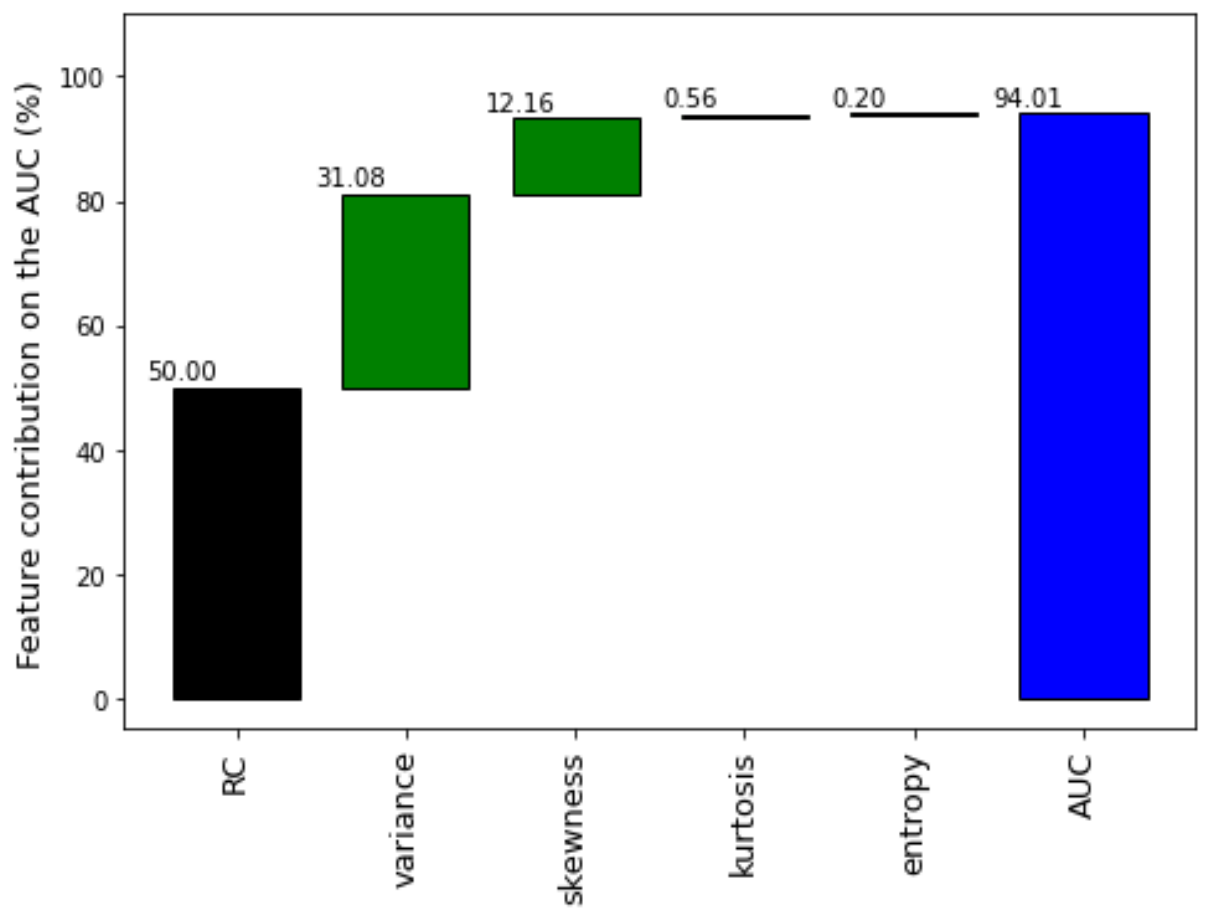}
\label{fig:robustness_stand_auc}}
\hfill
\caption{ROC and AUC for the banknotes dataset.}
\label{fig:robustness_stand}
\end{figure}

The ML model was trained using Gaussian Naive Bayes (implementation from scikit-learn), however, the proposed approach can be used with any other ML model. The dataset was split with 80\% for training and 20\% for testing purpose. The ROC curve obtained is shown in Figure~\ref{fig:robustness_stand_roc}, with th AUC value of 94.03\%. By applying the ShapAUC approach, we can interpret the contribution of each feature towards this AUC. Figure~\ref{fig:robustness_stand_auc} shows the obtained results (\textit{RC} represents the random classifier). We can see that the highest contribution is assigned to the \emph{variance} (31.08\%), while both \emph{kurtosis} and \emph{entropy} have very low contribution to the AUC (0.56\% and 0.20\%, respectively).

\begin{figure}[!htbp]
\centering
\subfloat[along with the ROC curve.]{\includegraphics[width=3.3in]{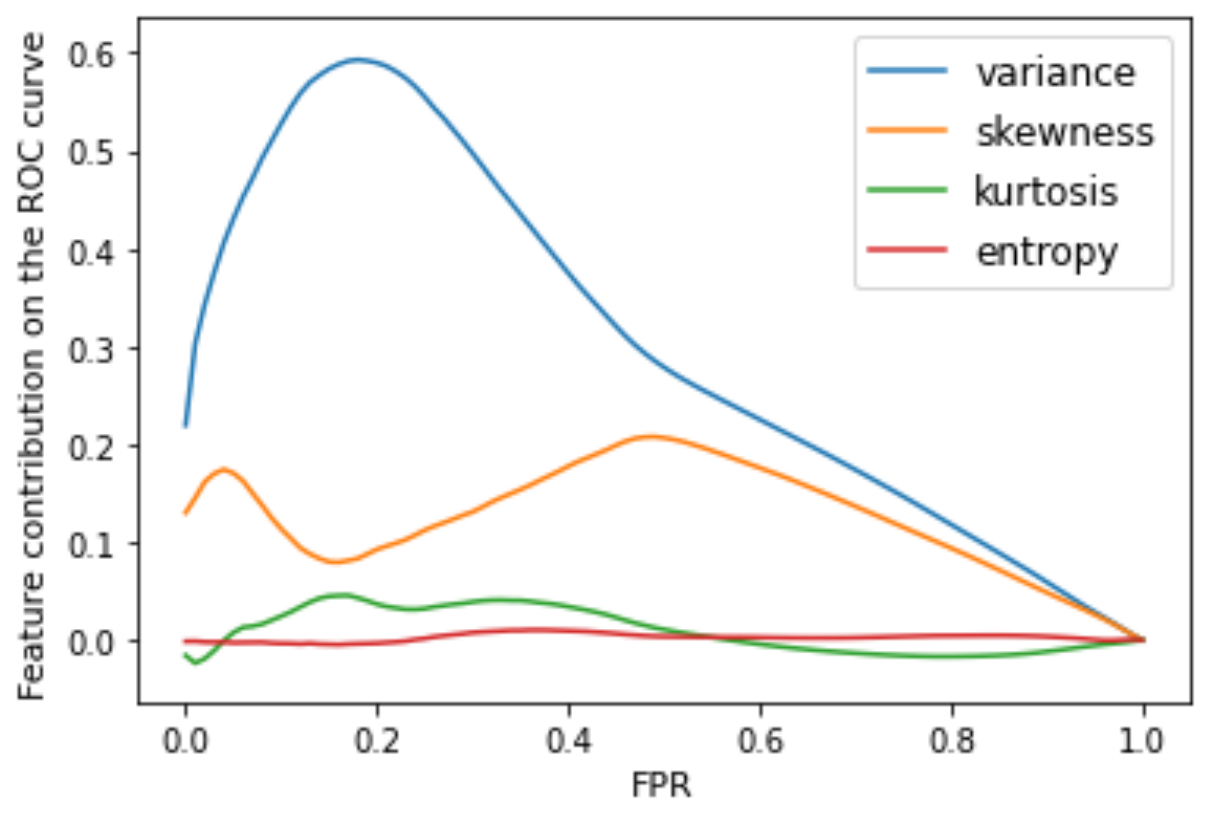}
\label{fig:robustness_stand_roc_explain_curves}}
\hfill
\subfloat[for a single slice (FPR of 20\%).]{\includegraphics[width=3.0in]{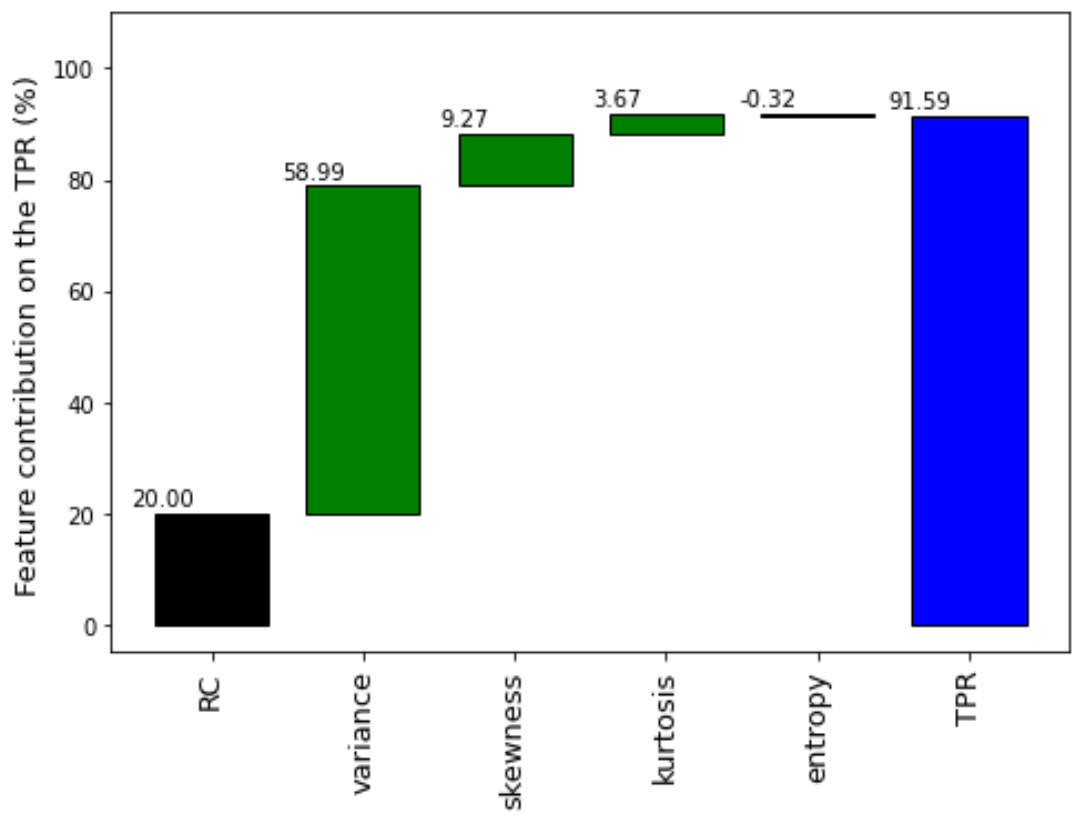}
\label{fig:robustness_stand_roc_explain_fpr20}}
\hfill
\caption{Feature contributions towards TPR values.}
\label{fig:robustness_stand_roc_explain}
\end{figure}

The results from the ShapROC approach are presented in Figure~\ref{fig:robustness_stand_roc_explain}. These results were generated with the interpolation strategy, however, it can be repeated for other strategies as well (see Section~\ref{subsec:trp_estimation}). As seen in Figure~\ref{fig:robustness_stand_roc_explain_curves}, \emph{variance} remains to be the attribute with most contribution towards the TPR values throughout the ROC curve. Both \emph{kurtosis} and \emph{entropy} have practically negligible contribution in the whole range of the FPR values. 

Figure~\ref{fig:robustness_stand_roc_explain_fpr20} shows the contributions for a FPR of 20\%, i.e. when moving from a TPR of 20\% (random classifier) to the actual 91.59\%. This waterfall plot can be considered as a slice view of Figure~\ref{fig:robustness_stand_roc_explain_curves} for FPR = 0.2. In this figure, an interesting observations is that the \emph{entropy} feature negatively contributes to the  TPR value. In other words, instead of improving the performance, \emph{entropy} is deteriorating the performance of the classifier at this point.

\subsection{On the use of different strategies for the TPR values estimation}
\label{subsec:trp_estimation}

\begin{figure*}[h!t]
\begin{centering}
\includegraphics[width=1.0\textwidth]{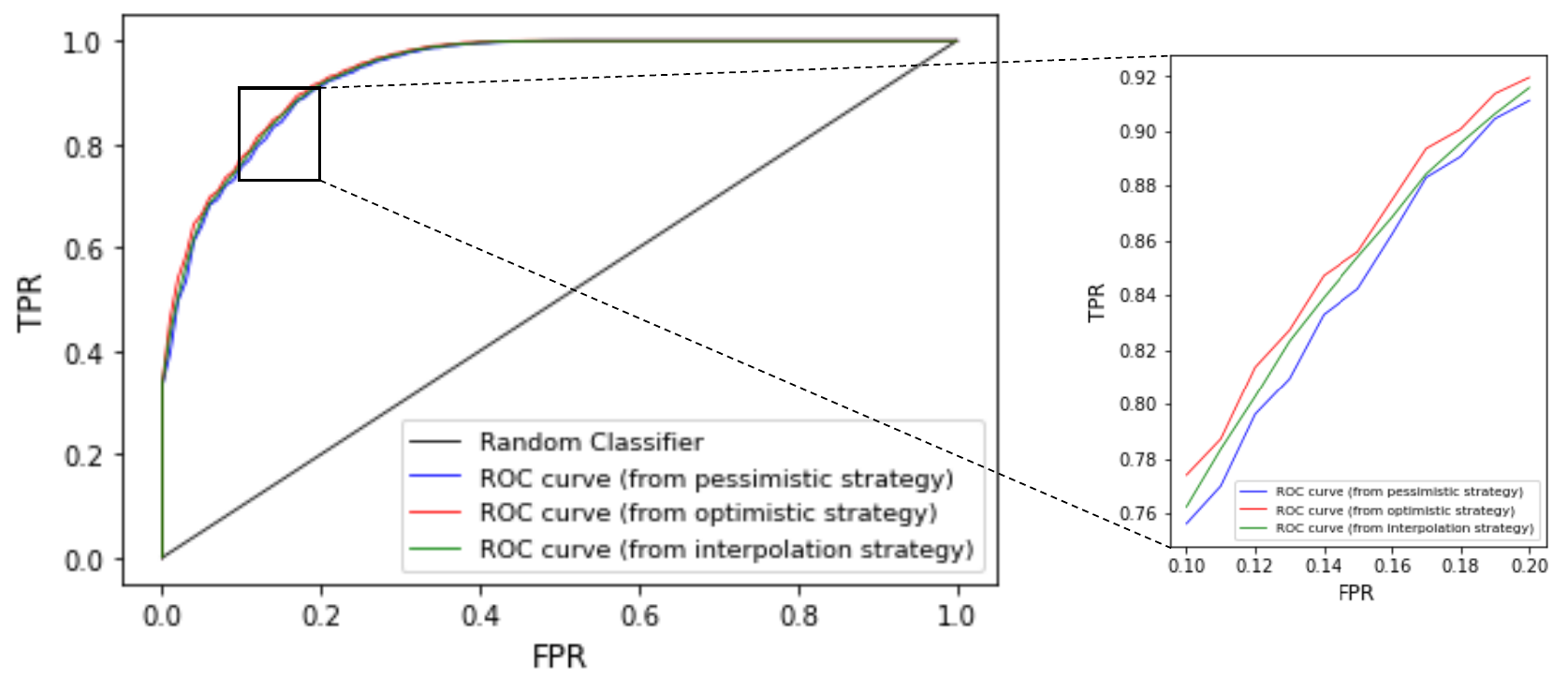} 
\par\end{centering}
\centering{}\caption{Estimated ROC curve for different strategies.
\label{fig:robustness_strategies}}
\end{figure*}

We mentioned in Section~\ref{subsec:shaproc_estimation} that different TPR values estimation strategies could be used in the proposed ShapROC approach. In this section, we discuss the different results that can be achieved by adopting the optimistic, pessimistic or  interpolation strategies. Figure~\ref{fig:robustness_strategies} presents the estimated ROC curve for these three strategies. As all curves are very similar, we show a magnified version of a selected piece of the ROC curves (to better visualise the differences). It can be seen that the optimistic strategy generates relatively higher values for TPRs and the pessimistic strategy has generated lower values. The interpolation strategy produces values between the other two strategies. 

\begin{figure}[!h]
\centering
\subfloat[Optimistic strategy.]{\includegraphics[width=2.8in]{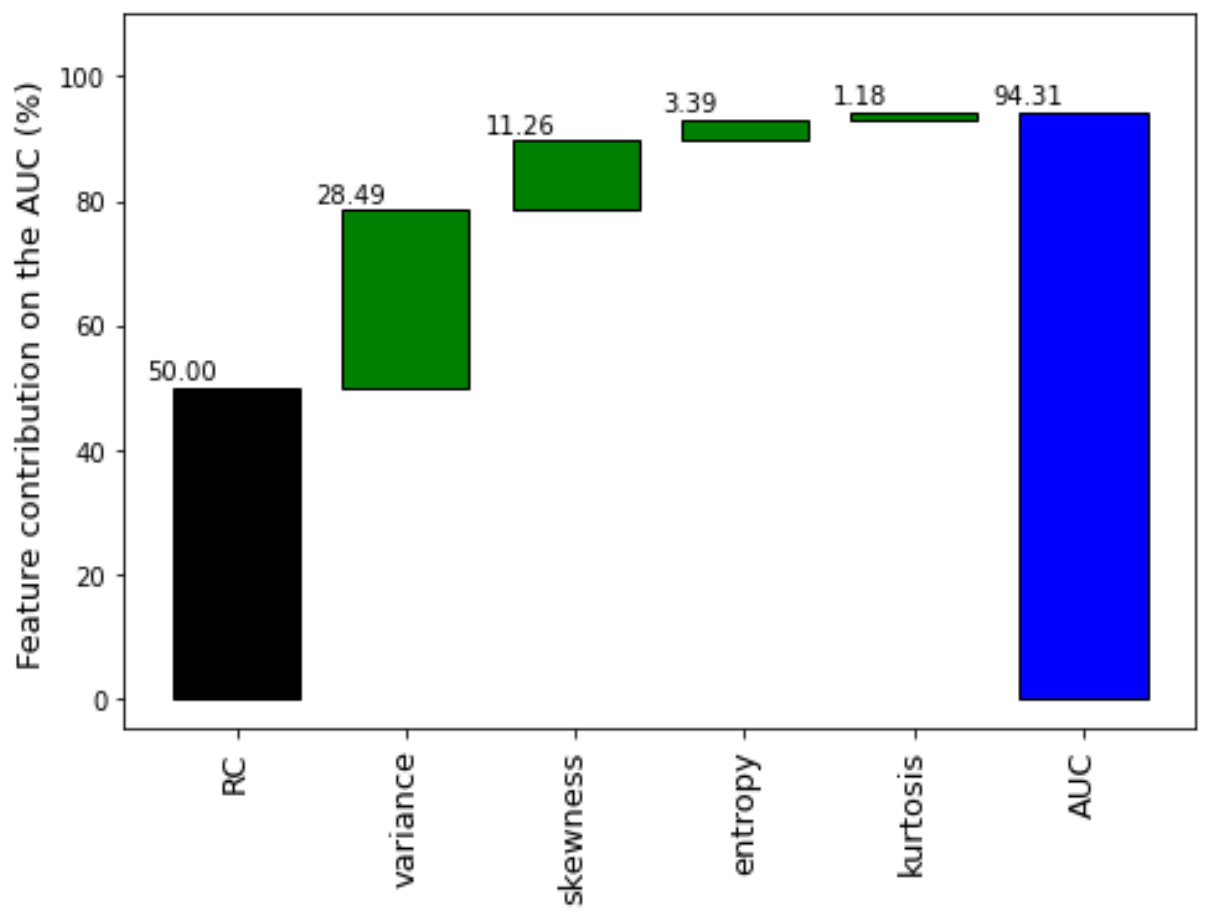}
\label{fig:robustness_auc_opt}}
\hfill
\subfloat[Optimistic strategy.]{\includegraphics[width=3.1in]{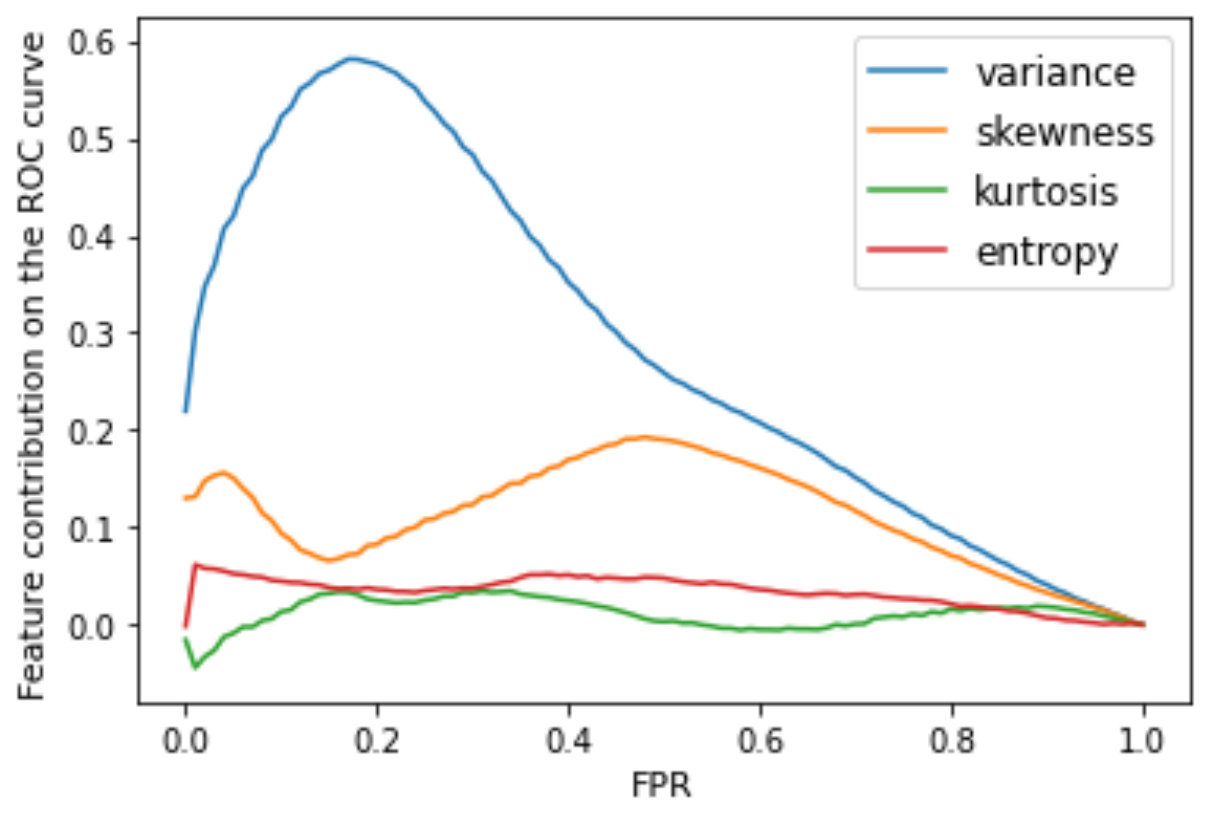}
\label{fig:robustness_roc_curves_opt}}
\hfill
\subfloat[Pessimistic strategy.]{\includegraphics[width=2.8in]{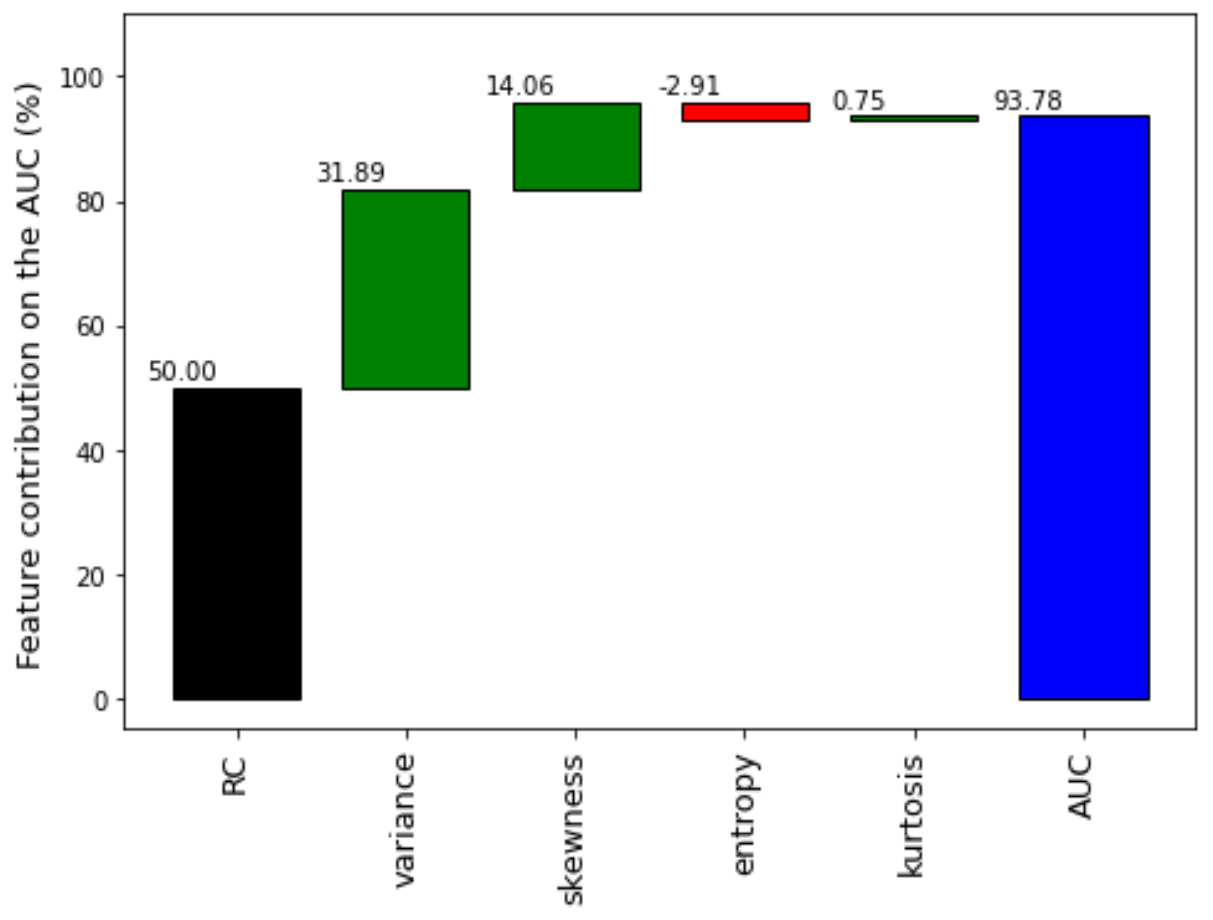}
\label{fig:robustness_auc_pes}}
\hfill
\subfloat[Pessimistic strategy.]{\includegraphics[width=3.1in]{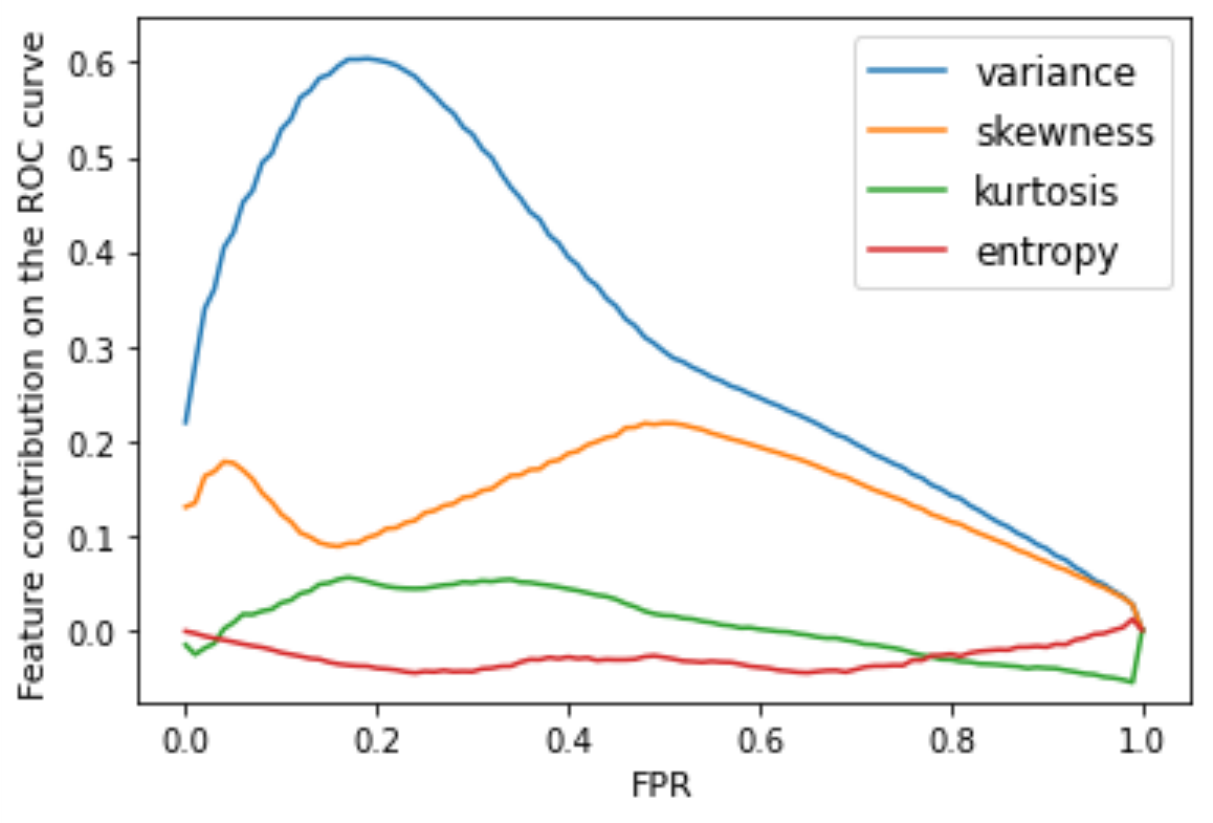}
\label{fig:robustness_roc_curves_pes}}
\hfill
\subfloat[Interpolation strategy.]{\includegraphics[width=2.8in]{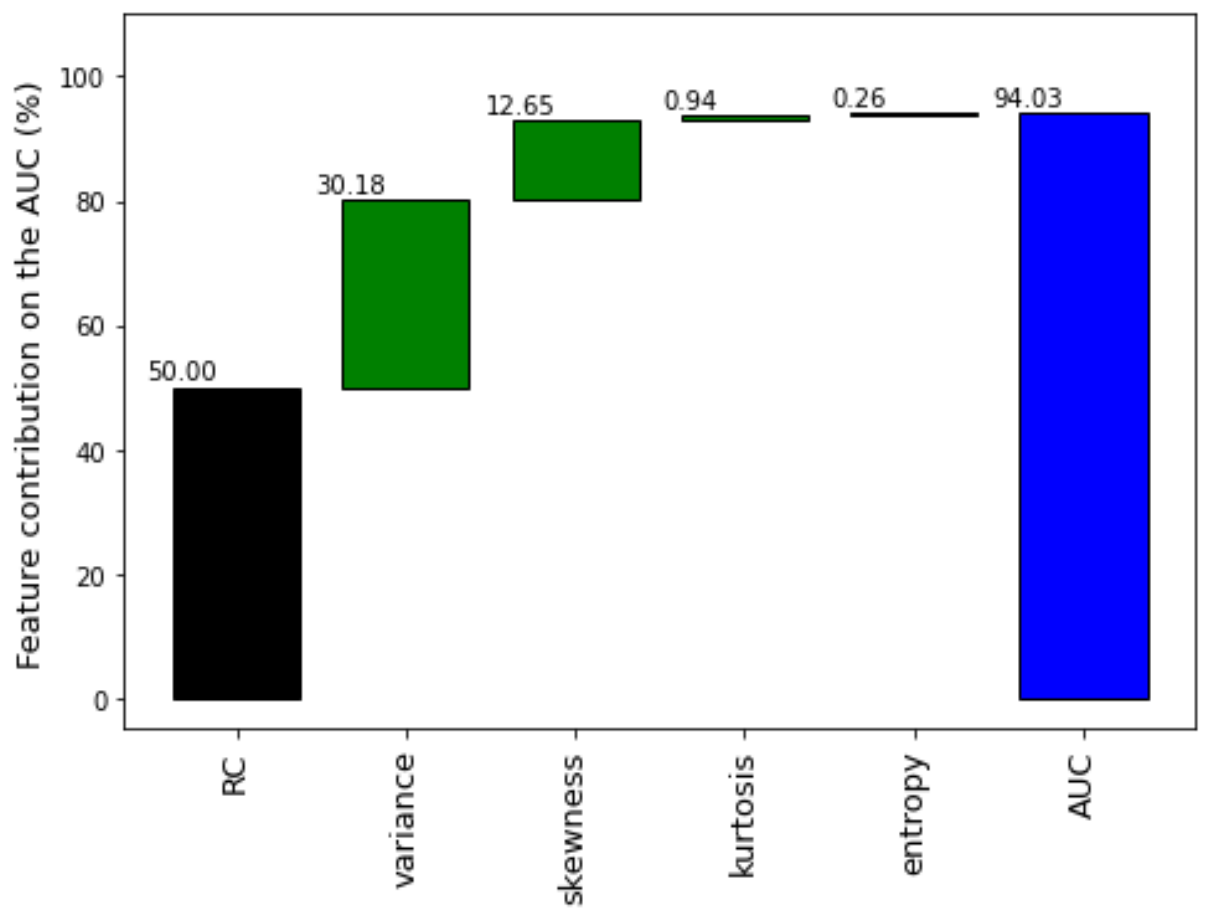}
\label{fig:robustness_auc_int}}
\hfill
\subfloat[Interpolation strategy.]{\includegraphics[width=3.1in]{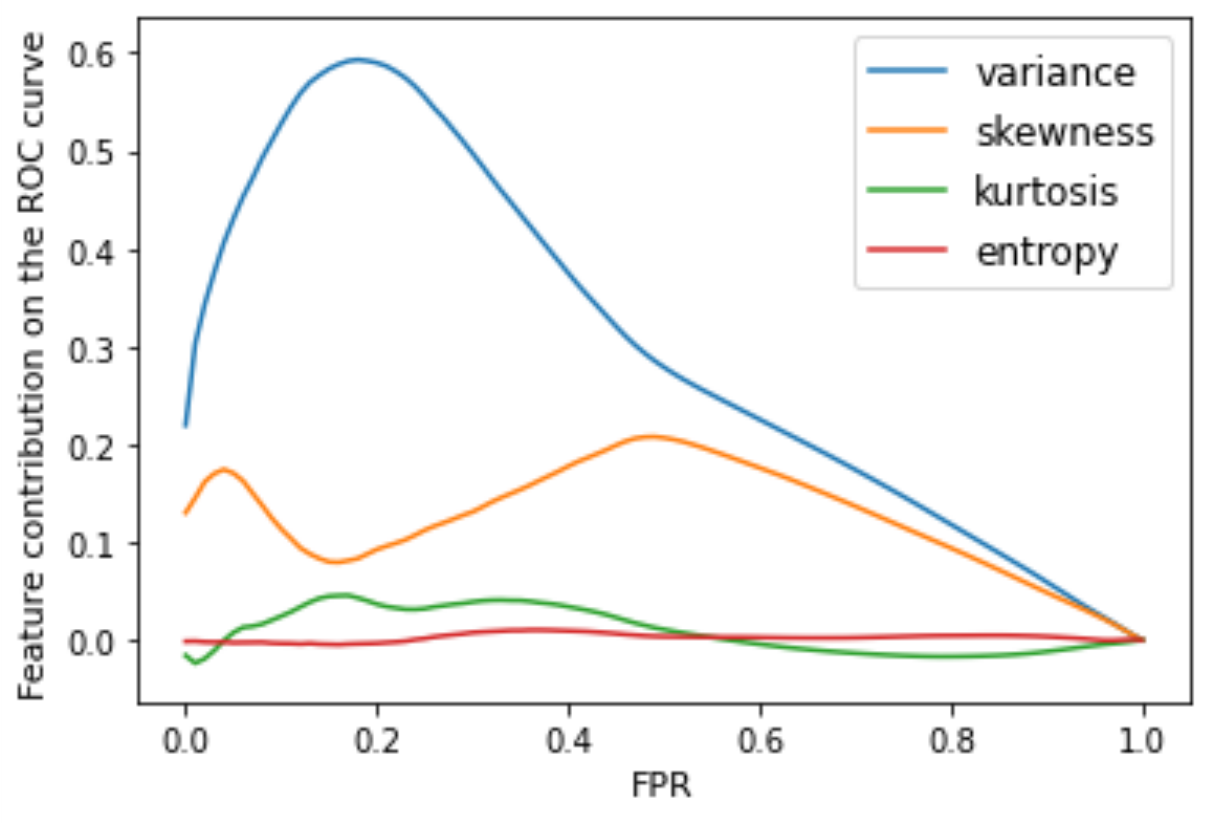}
\label{fig:robustness_roc_curves_int}}
\hfill
\caption{Comparison between the optimistic, pessimistic and interpolation strategies. Plots in the left: contributions towards the AUC. Plots in the right: contributions towards TPR values along with the ROC curve.}
\label{fig:robustness_strategies_curves}
\end{figure}

A comparison between the considered strategies is presented in Figure~\ref{fig:robustness_strategies_curves}. Although there are slight differences among the obtained results, in all cases, both \textit{variance} and \textit{skewness} are the two features with highest contributions towards the AUC and the TPR values (see Figures~\ref{fig:robustness_auc_opt},~\ref{fig:robustness_auc_pes} and~\ref{fig:robustness_auc_int}). 

One may also note in Figures~\ref{fig:robustness_roc_curves_opt},~\ref{fig:robustness_roc_curves_pes} and~\ref{fig:robustness_roc_curves_int} that the shapes of these features do not change a lot  with the strategy, although the shapes for \textit{kurtosis} and \textit{entropy} features change slightly. An interesting observation is that \textit{entropy} has a positive overall contribution towards AUC for the optimistic strategy while it has a negative contribution when using the pessimistic strategy. However, as both \textit{kurtosis} and \textit{entropy} have a very low contribution towards robustness, these differences can be considered negligible in terms of explainability.

As mentioned earlier, the aim here is not to propose or evaluate the best strategy, although this can be an area of future work. Without loss of generality, we will consider the interpolation strategy for the onward discussion in this paper.

\subsection{Visualising uncertainties in assessing robustness} \label{sec:uncertainties}

\begin{figure}[!h]
\centering
\subfloat[ROC curve.]{\includegraphics[width=3.1in]{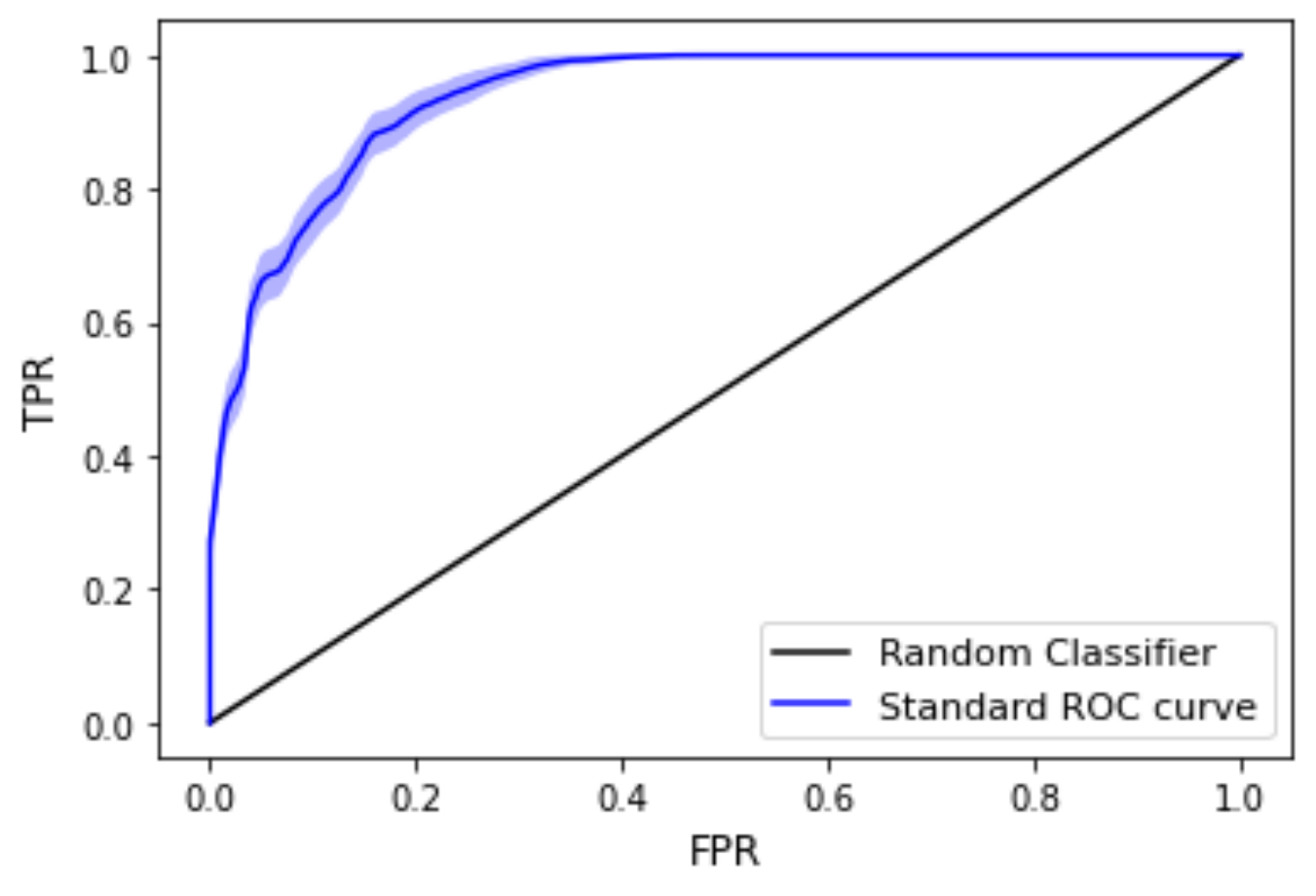}
\label{fig:robustness_roc_uncertain}}
\hfill
\subfloat[ShapAUC.]{\includegraphics[width=2.8in]{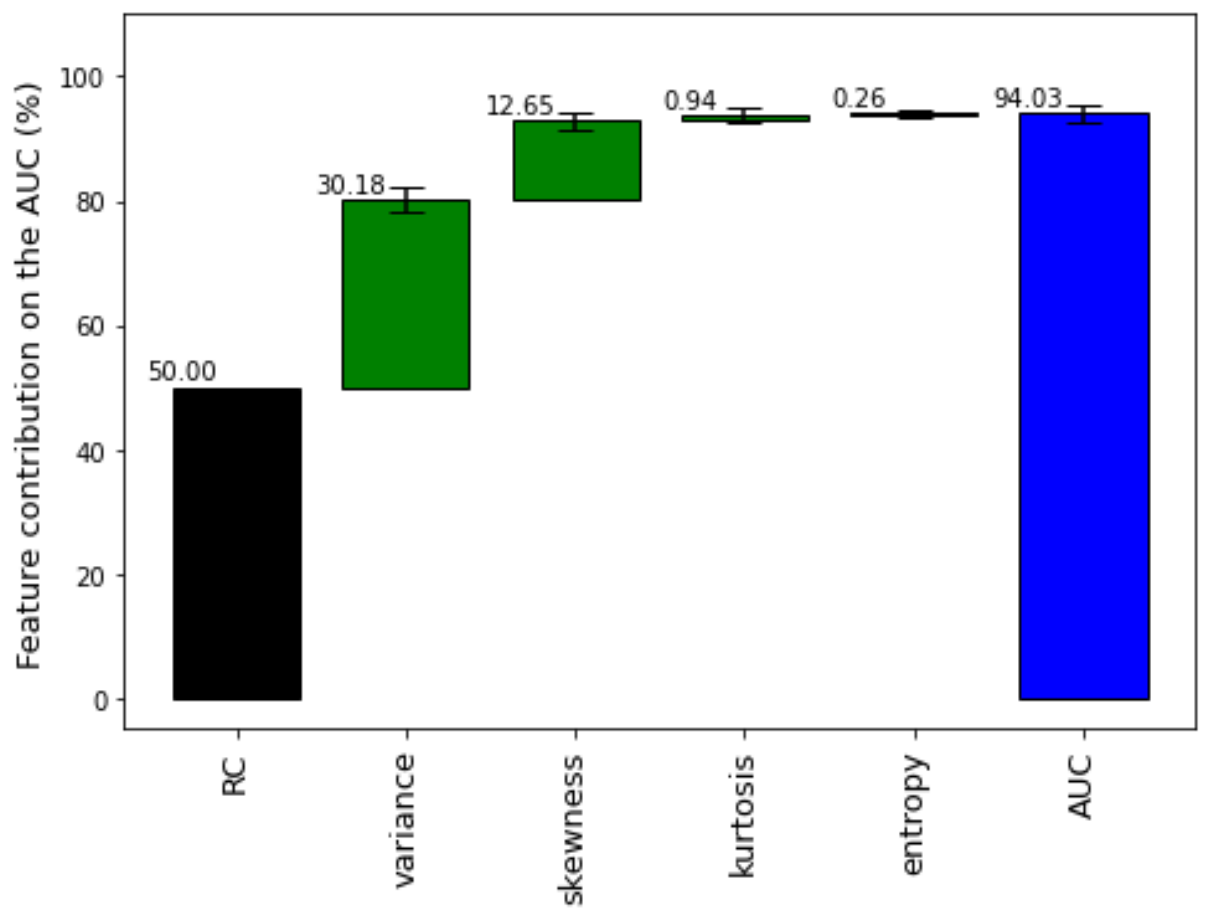}
\label{fig:robustness_auc_uncertain}}
\hfill
\caption{Uncertainties in ROC curve and ShapAUC.}
\label{fig:robustness_roc_auc_uncertain}
\end{figure}

A ML model should not be sensitive to a certain subset of the dataset selected for training (and/or testing), and therefore, it is a common practise to create multiple versions of training and testing datasets to assess the performance of ML models. These multiple versions can be created from original dataset by creating different combinations of subsets used for training and testing. The two most common approaches for performing these experiments are K-Fold Cross Validation \citep{refaeilzadeh2009cross} and Monte-Carlo Cross Validation \citep{xu2001monte}. Regardless of which approach we take, multiple experiments are involved that end up in multiple performance evaluations like AUC, ROC and PRC. For example, in Figure~\ref{fig:robustness_roc_uncertain}, we show the ROC curves obtained for the banknotes dataset using Monte-Carlo Cross Validation (using 100 iterations with uniformly distributed sampling). The crisp line in the middle shows the curve generated from expected values, and the shaded area around the curve shows the standard deviation in the values of these curves.

For each iteration, Shapley values can be used to estimate the contribution of each feature towards the AUC. These contributions may vary in each iteration, and therefore, we propose the dispersion in these values as a way of measuring uncertainty. Similarly, Figure~\ref{fig:robustness_auc_uncertain} explains the contribution of each feature towards the overall AUC where the bar height represents the expected contribution value while the whisker lines show standard deviation (i.e. uncertainty) in the contribution values.

\begin{figure}[!h]
\centering
\subfloat[Along with the ROC curve.]{\includegraphics[width=3.1in]{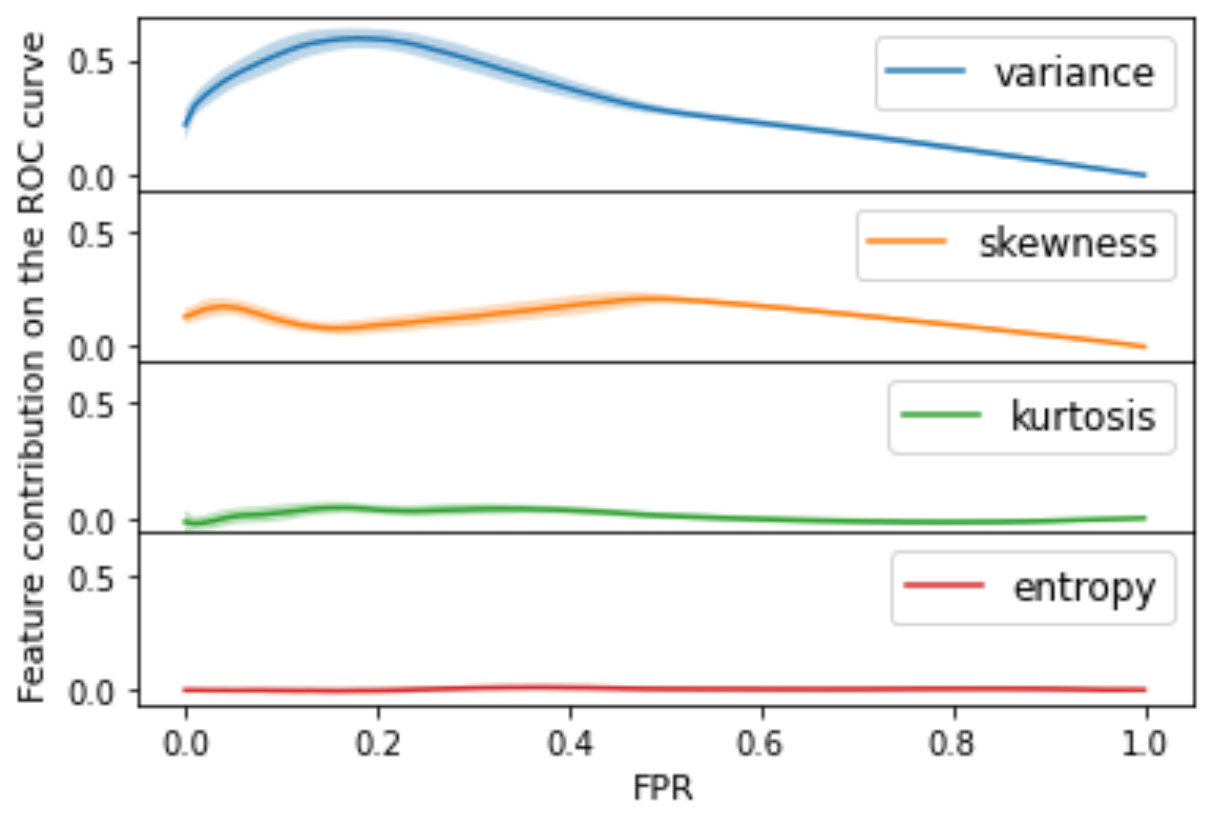}
\label{fig:robustness_roc_explain_uncertain}}
\hfill
\subfloat[For a single slice (FPR of 20\%).]{\includegraphics[width=2.8in]{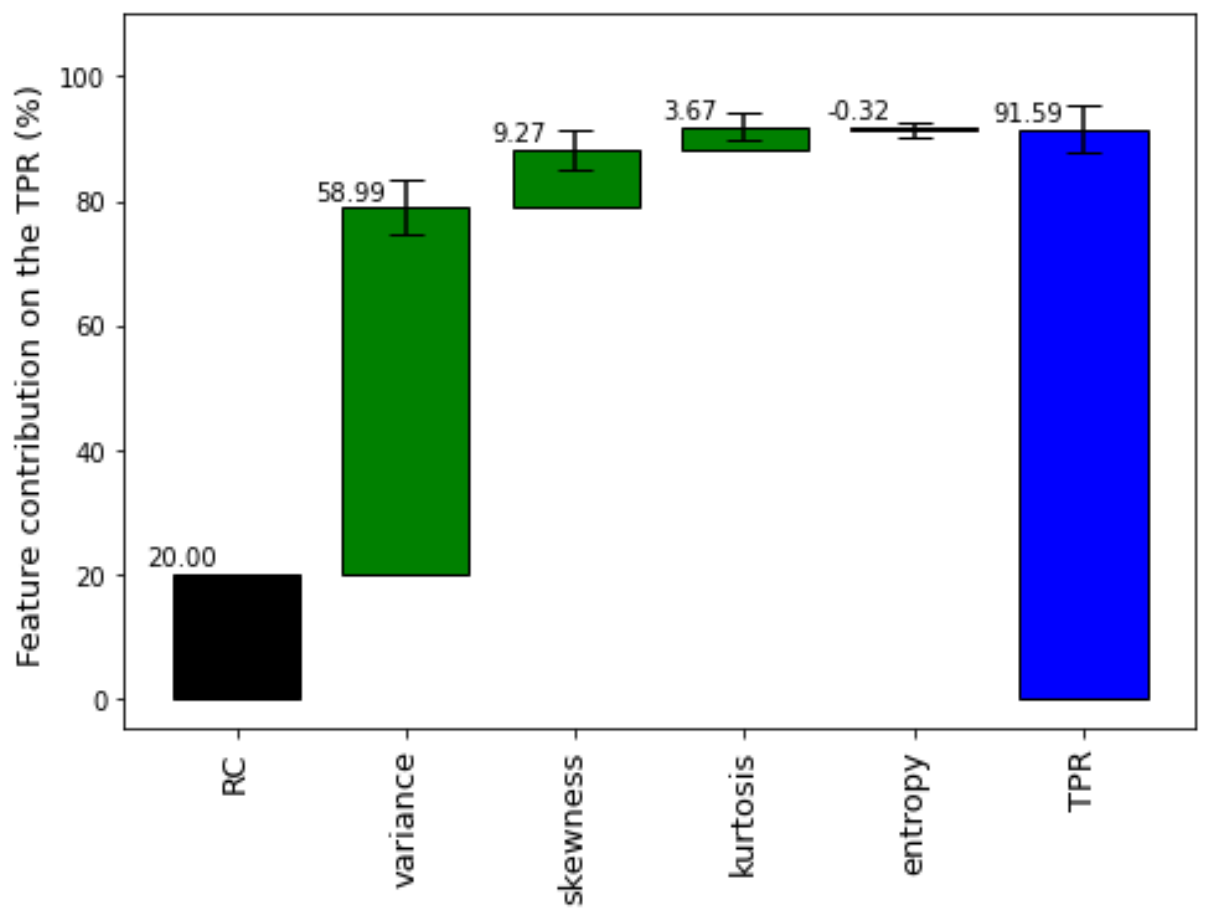}
\label{fig:robustness_roc_explain_fpr20_uncertain}}
\hfill
\caption{Uncertainties in ShapROC.}
\label{fig:robustness_uncertain}
\end{figure}

Extending this idea to the whole ROC curve, it is also possible to calculate these contributions of each attribute towards achieving a TPR value (for each FPR value in the ROC curve). This is demonstrated in Figure~\ref{fig:robustness_roc_explain_uncertain} where the four attributes of banknotes dataset are plotted separately. The figure shows expected contribution values as a solid line in the middle, while the shaded values depict the standard deviation in these contribution values. 

As a data analyst, one may need to investigate specific part of this plot, for example, focusing on the FPR value of 0.20, and investigating the contribution of each feature/attribute towards achieving the TPR value for FPR = 0.20. Figure~\ref{fig:robustness_roc_explain_fpr20_uncertain} shows such an example which is essentially a sliced view of the ROC curve shown in Figure~\ref{fig:robustness_roc_uncertain}. To summarise, both the measurement of robustness itself, as well as the uncertainty in measuring robustness can be explained with the help of Shapley values. 

\subsection{Analysing imbalanced classification datasets}

\begin{figure}[!h]
\centering
\subfloat[PRC.]{\includegraphics[width=3.1in]{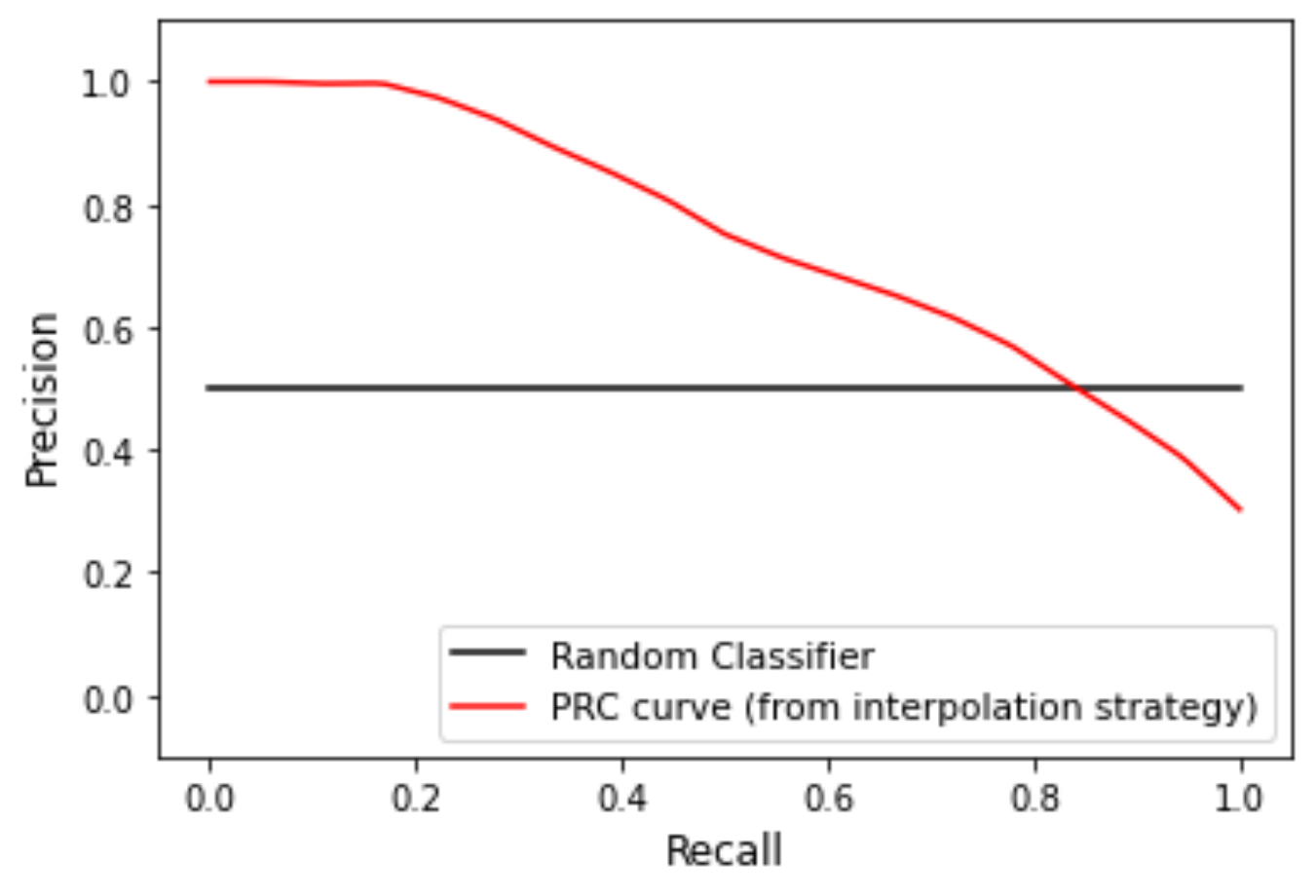}
\label{fig:robustness_prc_imbal}}
\hfill
\subfloat[ShapAUPRC.]{\includegraphics[width=2.8in]{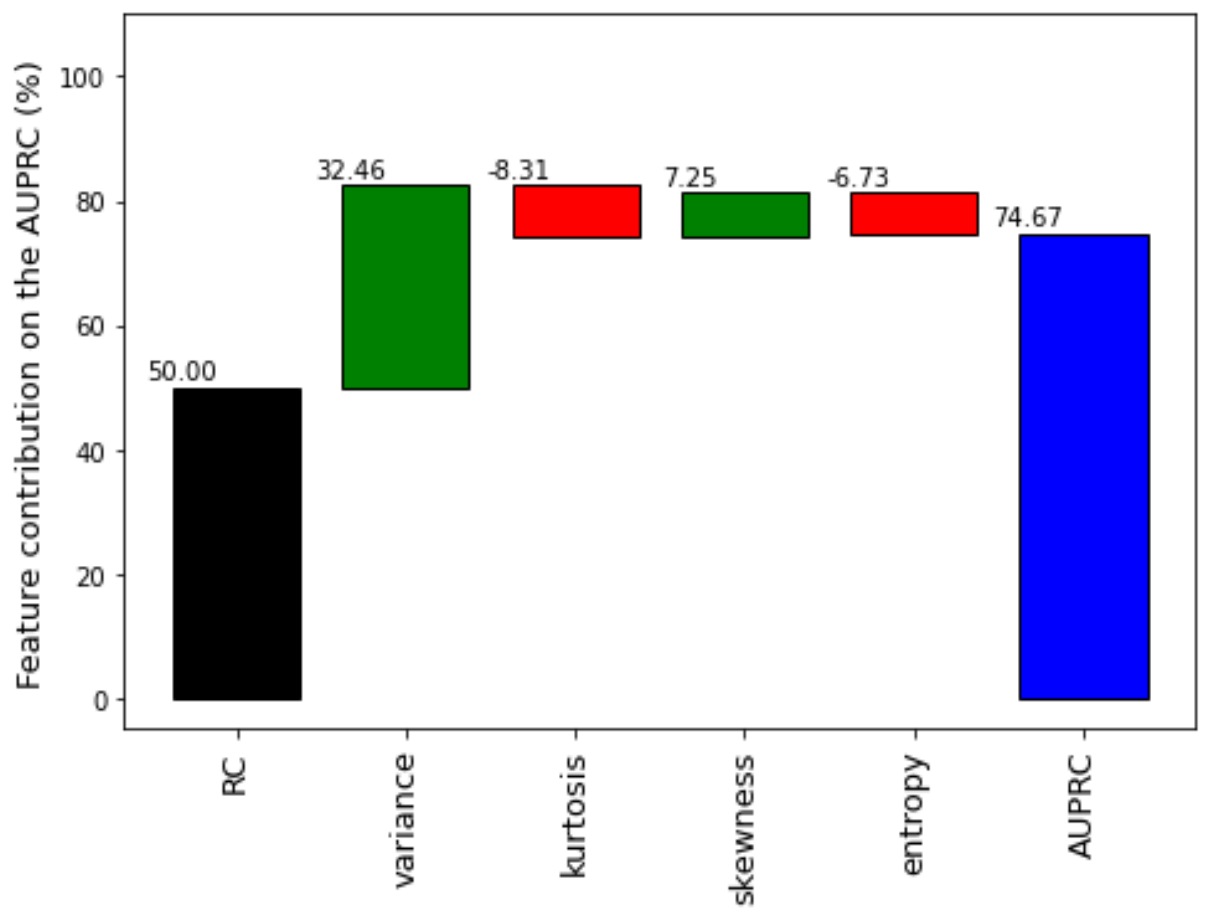}
\label{fig:robustness_aucpr_imbal}}
\hfill
\subfloat[ShapPRC.]{\includegraphics[width=3.2in]{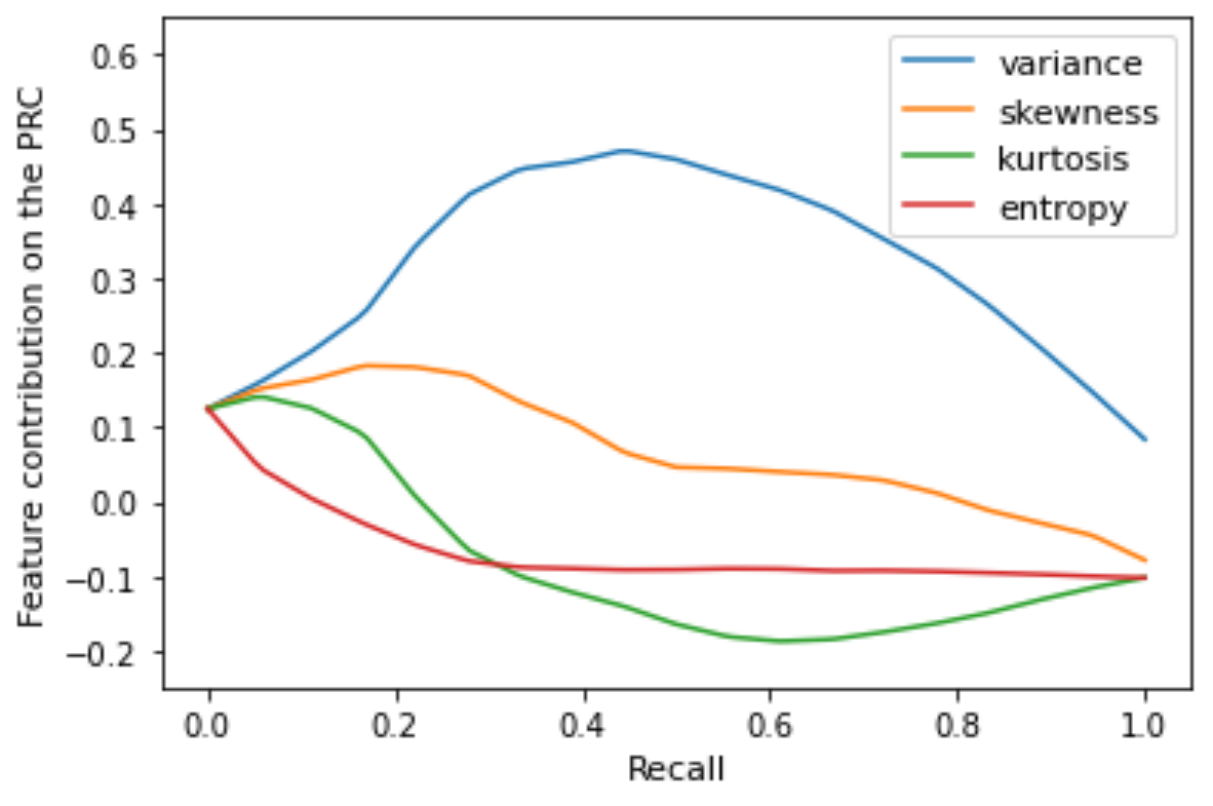}
\label{fig:robustness_prc_explain_imbal}}
\hfill
\caption{Application of ShapPRC and ShapAUPRC in imbalanced dataset.}
\label{fig:robustness_shapaucpr_imbal}
\end{figure}

As mentioned earlier, the use of ROC and AUC has been debated for imbalanced classification problems and the use of Precision-Recall Curve (PRC) is considered more appropriate \citep{cook2020consult}. We demonstrate the use of ShapPRC with the same illustrative example of banknotes. For demonstration purpose, we sub-sampled the banknotes data to synthetically create an imbalance of 90\%-10\% (with 10\% fake/forged notes). Figure \ref{fig:robustness_prc_imbal} shows the PRC for this derived dataset. The contributions of each feature towards achieving the AUPRC value is shown in Figure \ref{fig:robustness_aucpr_imbal}. The features of \textit{kurtosis} and \textit{entropy} can be seen to have negative contributions which implies that these features are decreasing the robustness (measured through PRC).

A more detailed picture can be seen in Figure \ref{fig:robustness_prc_explain_imbal}, where \textit{variance} and \textit{skewness} are contributing significantly higher than the other two. Please note that the aim of this experiment is not to compare PRC and ROC curves, as this is out of the scope of this paper. We demonstrate that both curves and their respective areas can be explained with the help of Shapley values.

\subsection{Using robustness analysis for feature engineering}

The gain or loss in the robustness by removing a specific feature is not necessarily equivalent to its marginal contribution. Instead, this is given by the difference between the payoffs with and without such a feature. However, as the Shapley value provides the marginal contribution when all possible coalitions are considered, it also serves as an indicative whether the presence (or absence) of a feature impacts the model robustness. Features whose contributions are insignificant could arguably be removed without affecting the robustness of the model.

\begin{figure}[!ht]
\centering
\subfloat[Removing \textit{kurtosis} and \textit{entropy}.]{\includegraphics[width=3.0in]{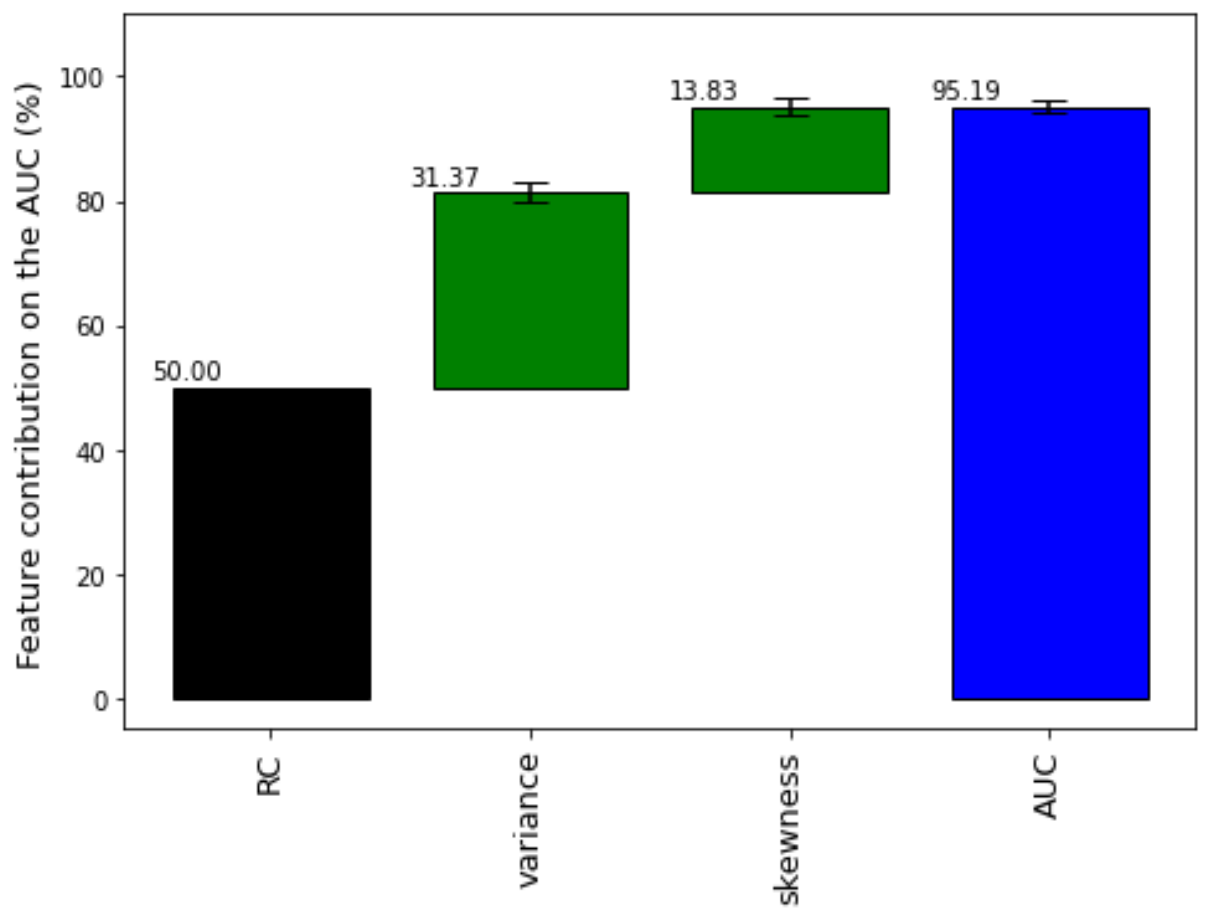}
\label{fig:robustness_auc_int_featB}}
\hfill
\subfloat[Removing \textit{variance}.]{\includegraphics[width=3.0in]{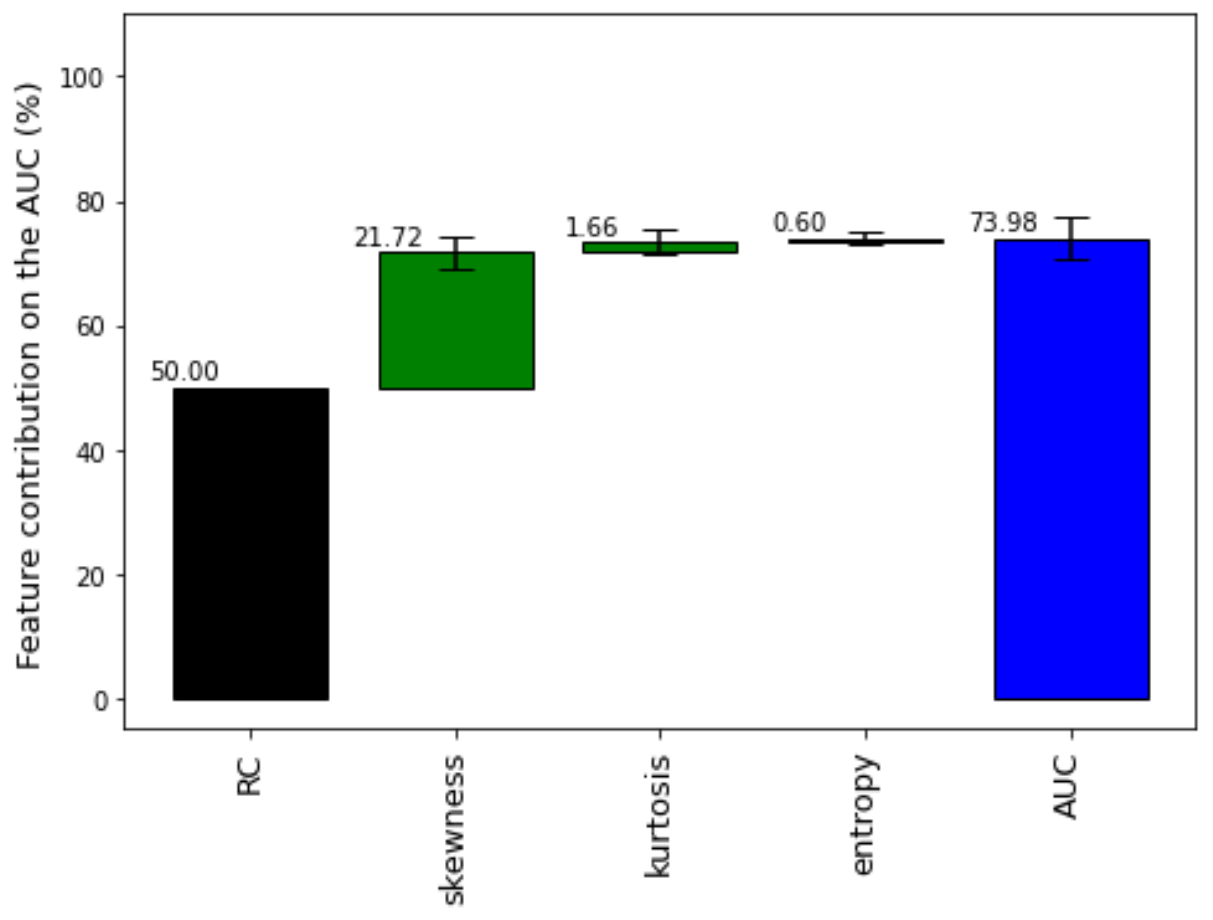}
\label{fig:robustness_auc_int_featG}}
\hfill
\caption{Application of ShapAUC as a feature selection method.}
\label{fig:robustness_featBG}
\end{figure}

Recall Figure~\ref{fig:robustness_auc_uncertain} for the illustrative example, it showed that both \textit{kurtosis} and \textit{entropy} have practically no contributions towards AUC while \textit{variance} has the highest contribution. By removing the \textit{kurtosis} and \textit{entropy} features, a slight improvement can be achieved on the model robustness. This is shown in Figure~\ref{fig:robustness_auc_int_featB} where the AUC increases slightly from 94.03\% to 95.19\%. However, if a more useful feature, like \textit{variance}, is removed from the dataset, the AUC may decrease significantly. This is demonstrated in Figure~\ref{fig:robustness_auc_int_featG} for the illustrative example where the AUC decrease from 94.03\% to 73.98\%, which is a detrimental change in robustness that might end up in making the model practically useless. 

\begin{figure}[!ht]
\centering
\subfloat[Contributions towards AUC.]{\includegraphics[width=2.7in]{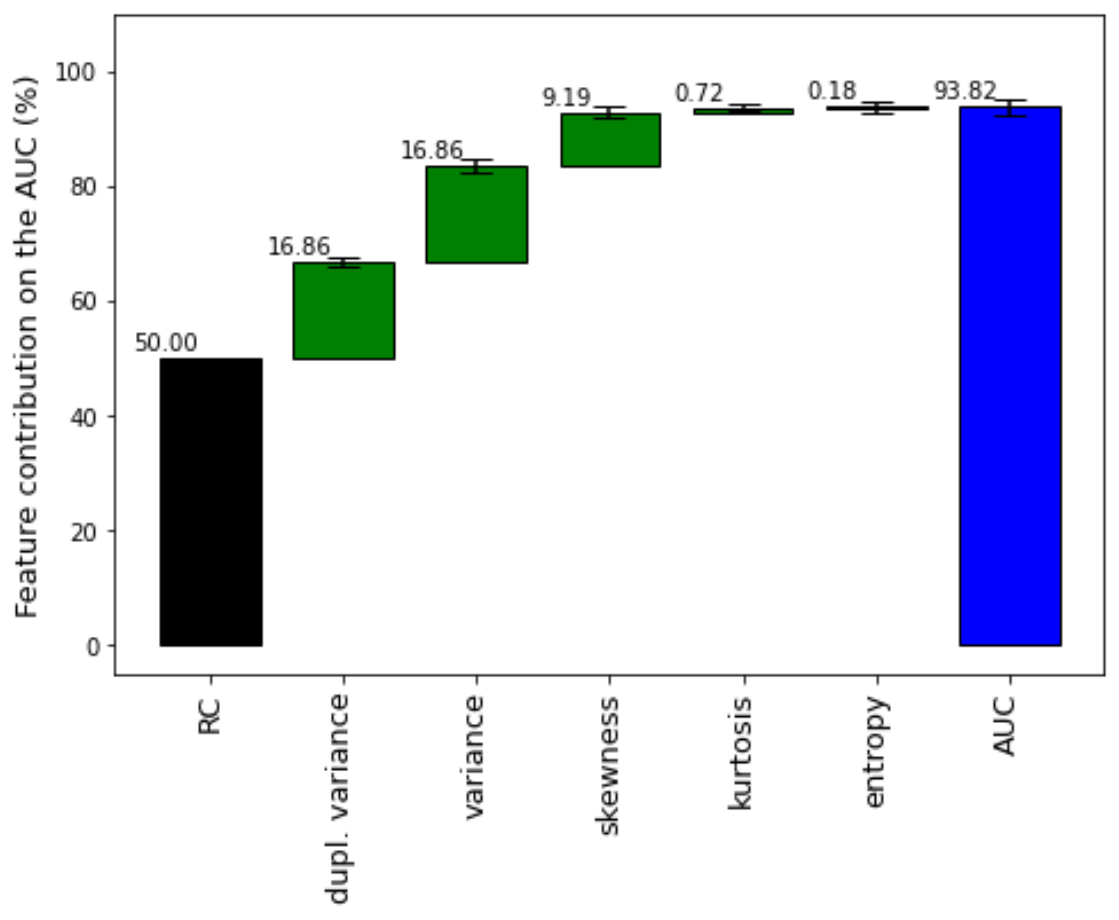}
\label{fig:robustness_auc_int_dupl}}
\hfill
\subfloat[Contributions along with the ROC curve.]{\includegraphics[width=3.2in]{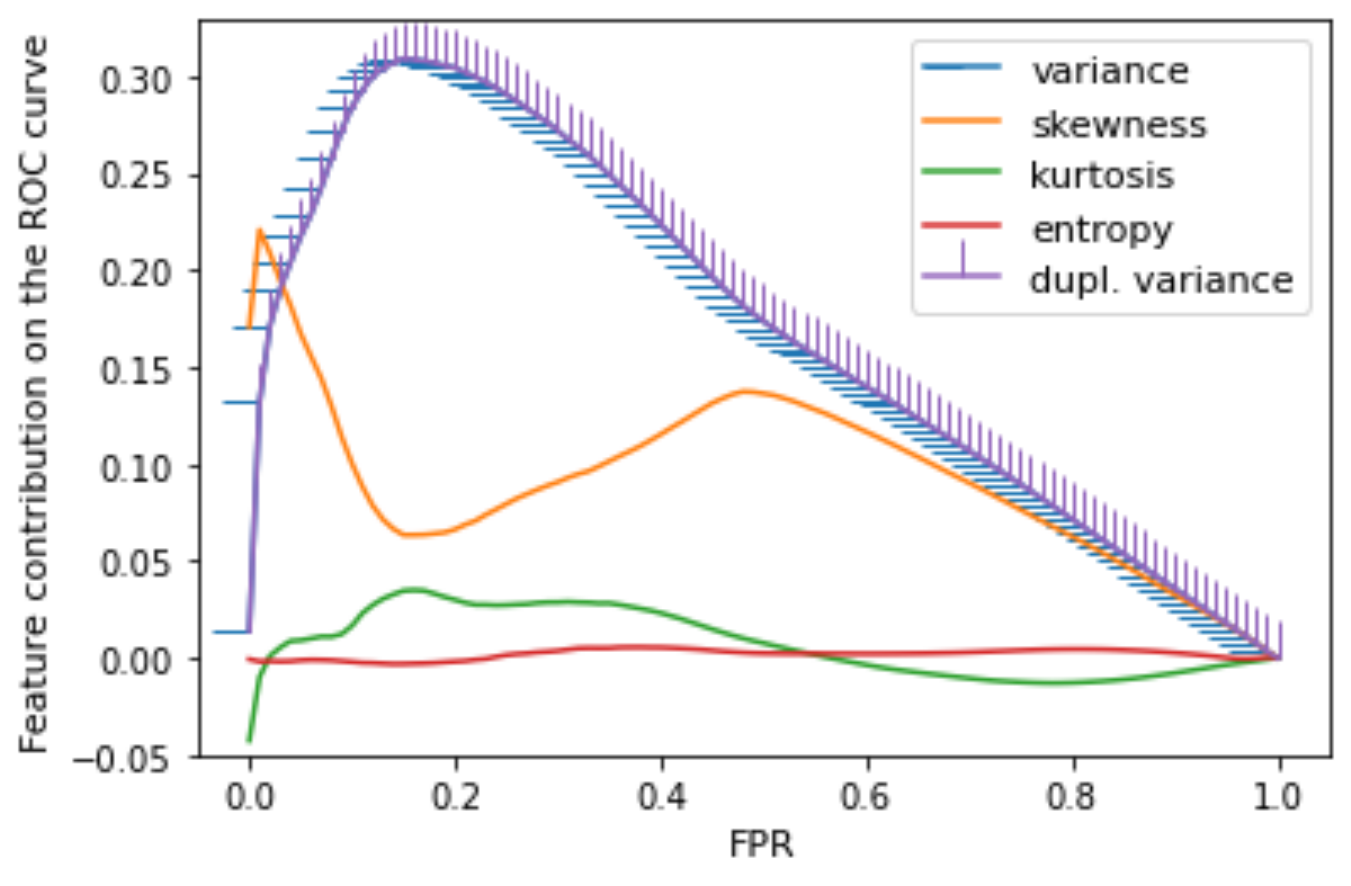}
\label{fig:robustness_stand_roc_explain_curves_dupl}}
\hfill
\caption{Evaluating the inclusion of a duplicate \textit{variance}.}
\label{fig:robustness_dupl}
\end{figure}

Also, to investigate the impact of having duplicate feature, we create a novel feature which is a copy of \textit{variance}. Figure~\ref{fig:robustness_dupl} shows the contributions of this duplicate variable towards the AUC and the ROC curve. In Figure~\ref{fig:robustness_auc_int_dupl}, we see that both \textit{variance} and the novel feature (represented by \textit{dupl. variance}) contributes equally towards the AUC (16.86\%). This is attested in Figure~\ref{fig:robustness_stand_roc_explain_curves_dupl} where the contributions of the duplicate variance are identical to the original variance curve. The figure uses vertical and horizontal markers on these two curves to highlight their overlap. As the novel feature is a duplicate of \textit{variance}, the payoffs $\upsilon \left(A \cup \left\{variance \right\} \right) = \upsilon \left(A \cup \left\{\text{\textit{dupl. variance}} \right\} \right)$ and, therefore, $\phi_{variance} = \phi_{\text{\textit{dupl. variance}}}$ (see the Symmetry property in Section~\ref{subsec:shapley_prop}).

\begin{figure}[!ht]
\centering
\subfloat[PRCs.]{\includegraphics[width=3.1in]{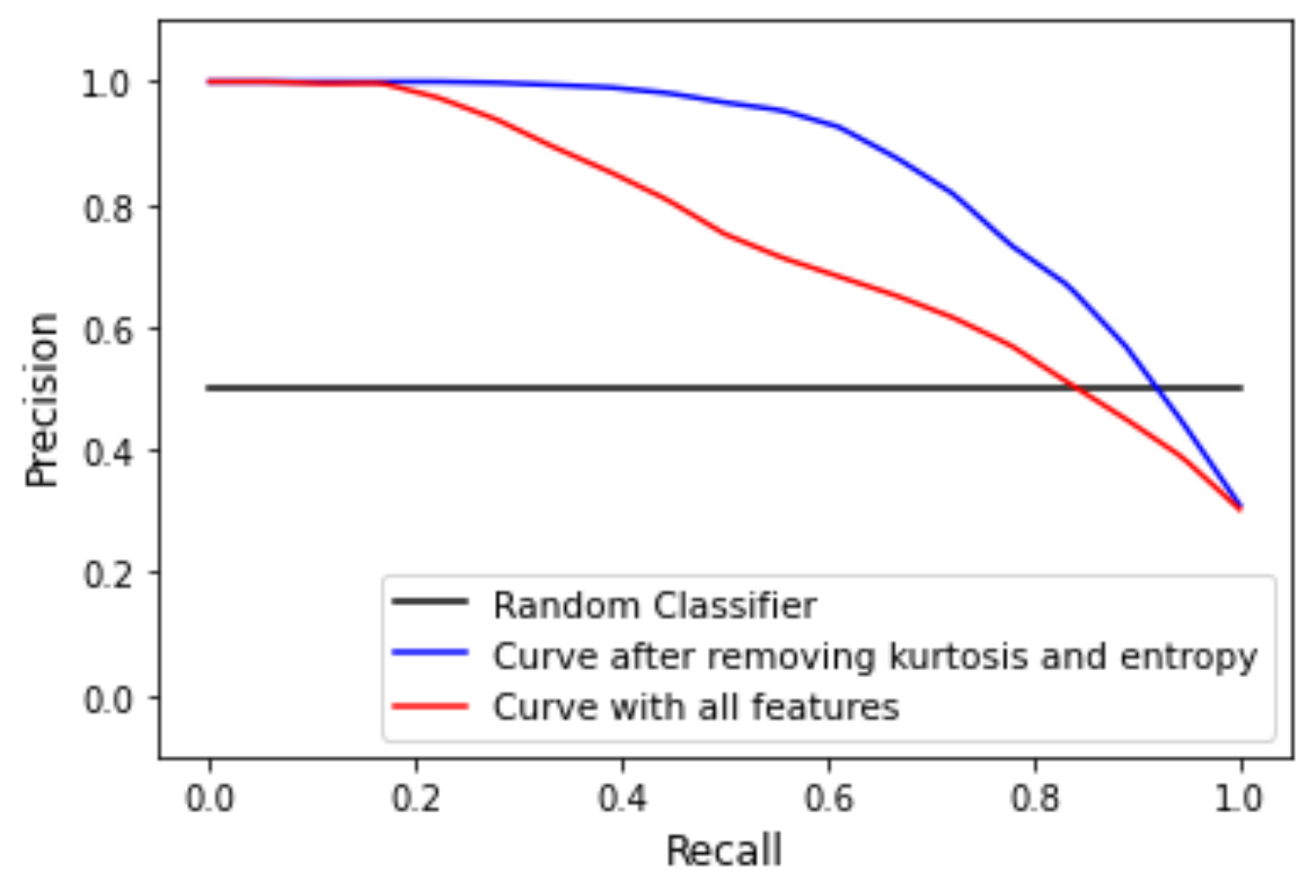}
\label{fig:robustness_prc_imbal_featB}}
\hfill
\subfloat[AUPRCs.]{\includegraphics[width=2.8in]{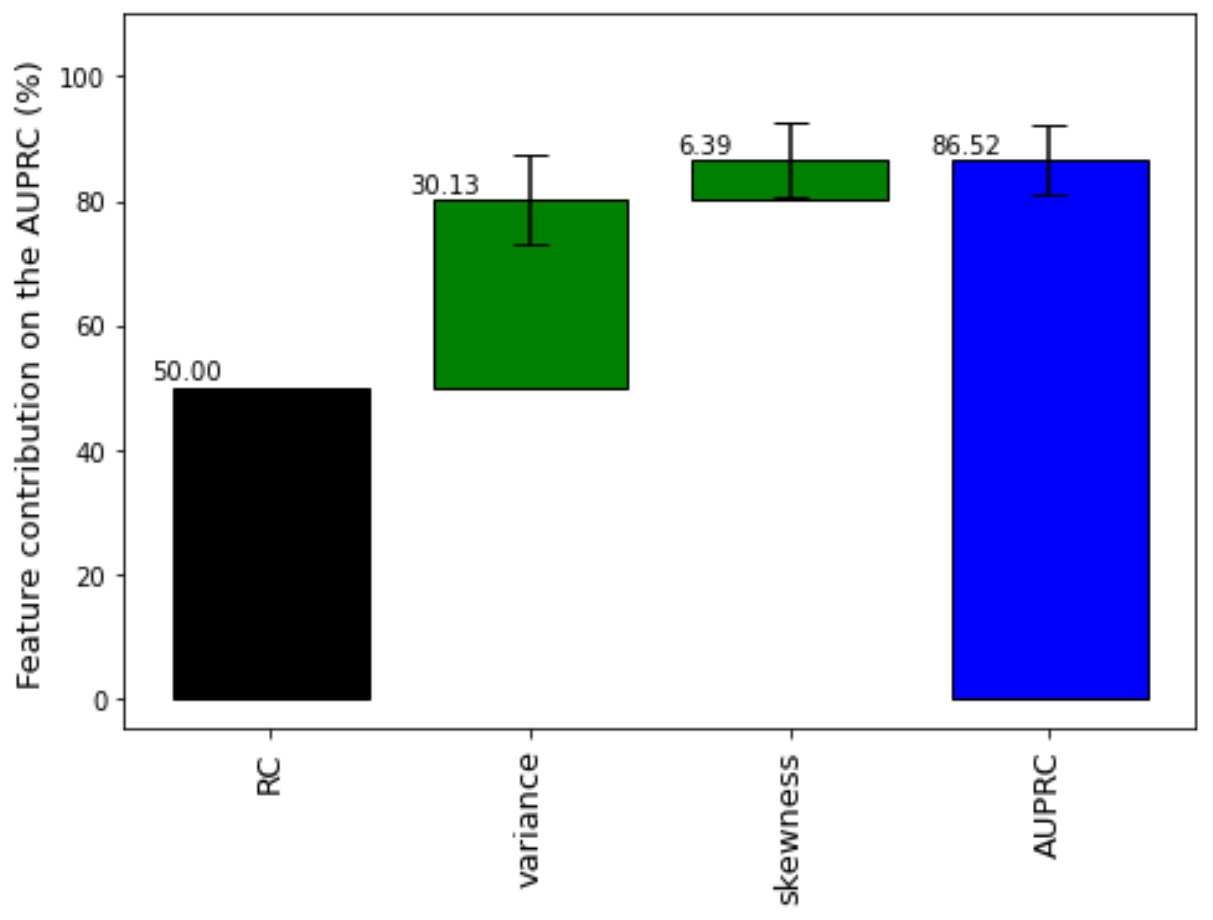}
\label{fig:robustness_aucpr_imbal_featB}}
\hfill
\caption{Application of ShapAUPRC as a feature selection method.}
\label{fig:robustness_imbal_featB}
\end{figure}

Note that the use of PRCs is preferred for assessing robustness in case of having imbalanced dataset. Recall Figure~\ref{fig:robustness_aucpr_imbal} where \textit{kurtosis} and \textit{entropy} both contributed negatively towards the AUPRC. Therefore, an obvious recommendation would be to remove these two features from the model. Figure~\ref{fig:robustness_aucpr_imbal_featB} shows the results after removing these two features. Clearly, the performance of the classifier has improved from 74.67\% to 86.52\%. This is a significant improvement in a sense that the model has improved by more than 15\%. 

This demonstrates the usefulness of our proposed approach in selecting or removing features, which is a critical step in machine learning applications. Appendix-I further illustrates the use of our proposed approach on the datasets for Red Wine Quality~\citep{Cortez2009}, Rice~\citep{Cinar2019}, and Pima Indians Diabetes~\citep{Smith1988} as well.

\section{Conclusions}
\label{sec:conclu}
We proposed techniques that can be used to explain the contribution of each feature towards the robustness of ML models. For explaining the area under the ROC curve, we propose to use 0.50 as the baseline value as any random classifier can achieve this value without help from any useful feature. Then, we propose to estimate the contribution of features towards adding robustness with the help of Shapley values. We also propose to explain each point at the ROC curve and therefore creating a decomposition of an overall curve into a set of individual curves (for each feature). As the use of PRC is considered more appropriate for imbalanced datasets, we also extended the idea to use Shapley values for explaining the PRC and the AUPRC. Explaining the robustness of classifiers can help analysts in auditing various features in their models and to revise their performance tuning parameters accordingly.

We demonstrate the use of our proposed approaches in feature selection. Based on the estimated Shapley values, it is possible to spot a feature that contributes negatively, and therefore, can be removed from the model. Also, this can help us identify features that should not be removed from the model due to their critical contributions towards robustness. In addition, it is also possible to identify a feature having insignificant contribution to the model's robustness, and therefore, removing such feature might help increasing the overall computational efficiency. 

\subsection{Limitations and future work}
As mentioned in Section \ref{subsec:shaproc_estimation}, to explain a ROC curve, we generate multiple curves where each curve represents one of the coalitions among the players (i.e. features). Comparing these curves is not a straightforward task due to the fact that each curve has a different set of FPR/TPR values which does not necessarily align with other curves. For this, we proposed the three possible strategies for estimating these values: optimistic, pessimistic and interpolation strategies. However, one may argue that these strategies are sub-optimal, and better strategies are possible. Therefore, we consider this an area of further research.

\begin{figure*}[h!t]
\begin{centering}
\includegraphics[width=0.85\textwidth]{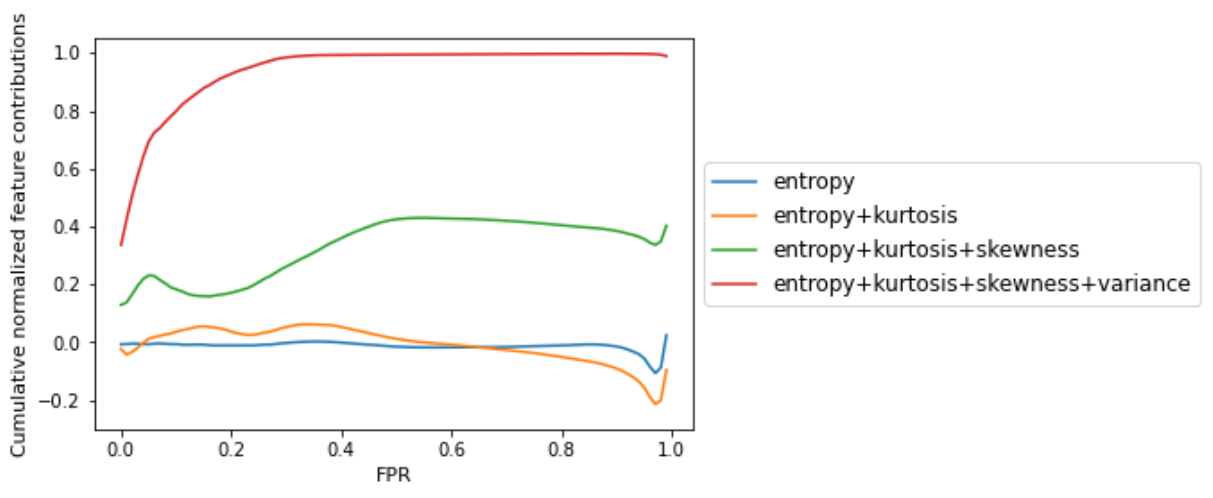} 
\par\end{centering}
\centering{}\caption{Example of relative feature contributions visualisation.
\label{fig:robustness_relat_contr}}
\end{figure*}

The visualisation of ShapROC (as shown in Figure \ref{fig:robustness_stand_roc_explain_curves}) can assist data analysts in assessing the contribution of each feature across a range of FPR values. However, as we move on this curve from left to right, the contribution of random classifier dominates the contribution of features. This is visible in the figure where all individual curves are approaching zero, and therefore, the relative importance of these features cannot be inspected visually. An analyst might be interested in assessing these contributions of features in a relative sense, and therefore, a normalised version of this plot might be more useful in such case. We show an example of normalisation in Figure~\ref{fig:robustness_relat_contr} which might be more useful when comparing the relative contributions of features towards achieving the TPR values. The same can be applied to ShapPRC plots as well. We consider this another area of future work that can help data analysts and researchers.

We demonstrated the use of Shapley values to explain the contribution of each feature towards the robustness of classifiers. However, this idea can be extended further to assess the interaction among the features. In some cases, it might be important to estimate the contribution of some combinations of features instead of treating them as standalone/independent features. We consider this another important area of future work with practical implications.

Finally, it is possible to further investigate datasets from different application domains using the proposed approaches, which is considered to be another area of future research. 

\section*{Acknowledgements}

The authors would like to thank the grants \#2021/11086-0, \#2020/10572-5 and \#2020/09838-0, S\~{a}o Paulo Research Foundation (FAPESP), for the financial support. The authors would also like to thank the Department of Management in Leeds University Business School for providing resources to conduct this research.

\bibliography{references}
\bibliographystyle{elsarticle-harv}

\section*{Appendix-I}

The following three Figures demonstrate the contribution of each feature towards the AUC for the Red Wine Quality~\citep{Cortez2009}, Rice~\citep{Cinar2019}, and Pima Indians Diabetes~\citep{Smith1988} datasets. By removing features  \textit{pH} and \textit{free sulfur dioxide} in the Wine dataset, the overall AUC have improved from 86.95\% to 87.15\%. For the Rice dataset, by removing feature \textit{Extent}, the overall AUC practically remained the same (from 95.27\% to 95.29\%). In the Diabetes dataset, by removing feature \textit{Glucose}, which has the highest contribution towards AUC, the robustness decreased from 79.82\% to 72.59\%.

\begin{figure}[!ht]
\centering
\subfloat[With all features.]{\includegraphics[width=3.0in]{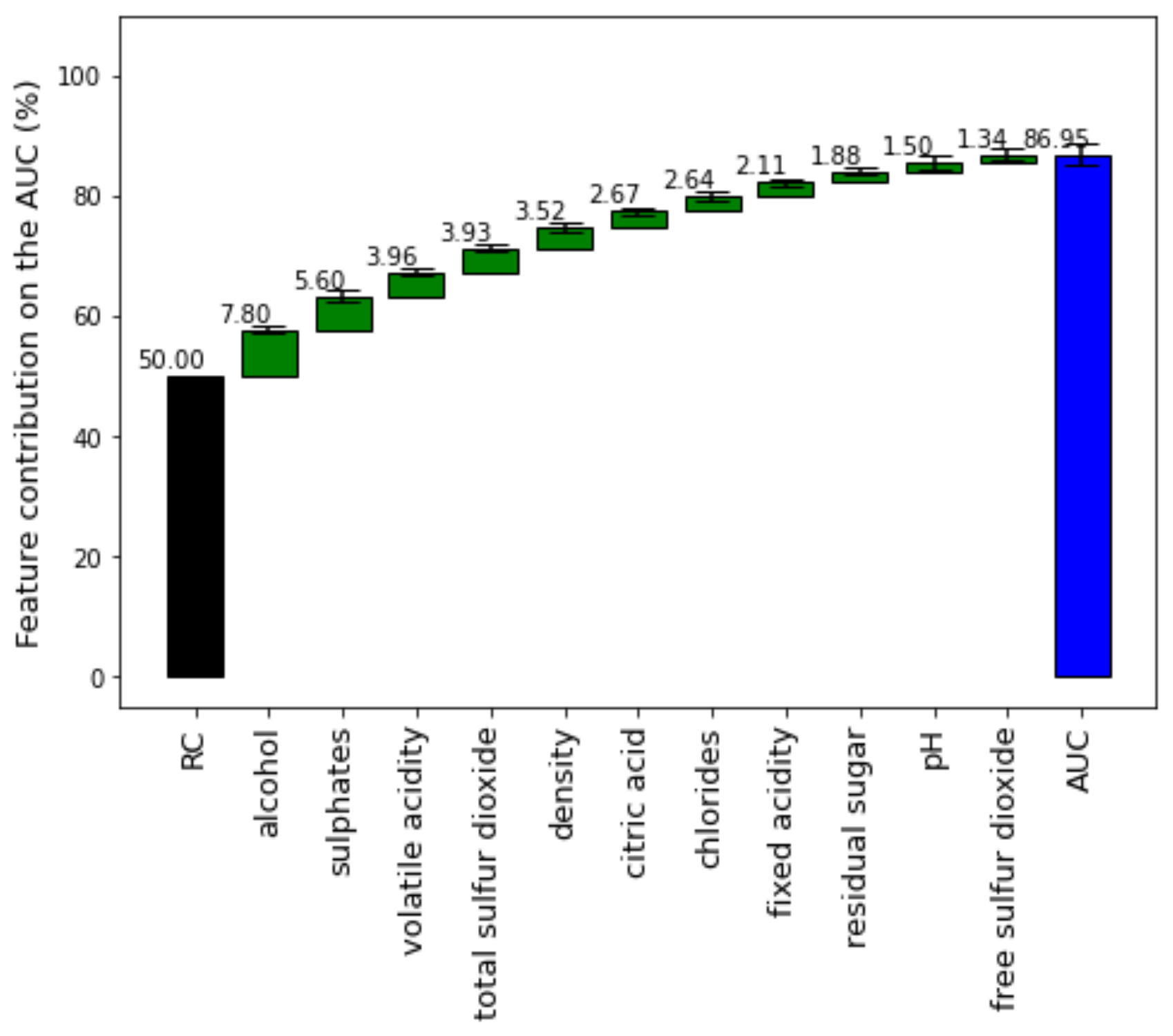}
\label{fig:robustness_auc_int_wine}}
\hfill
\subfloat[Removing \textit{pH} and \textit{free sulfur dioxide}.]{\includegraphics[width=3.0in]{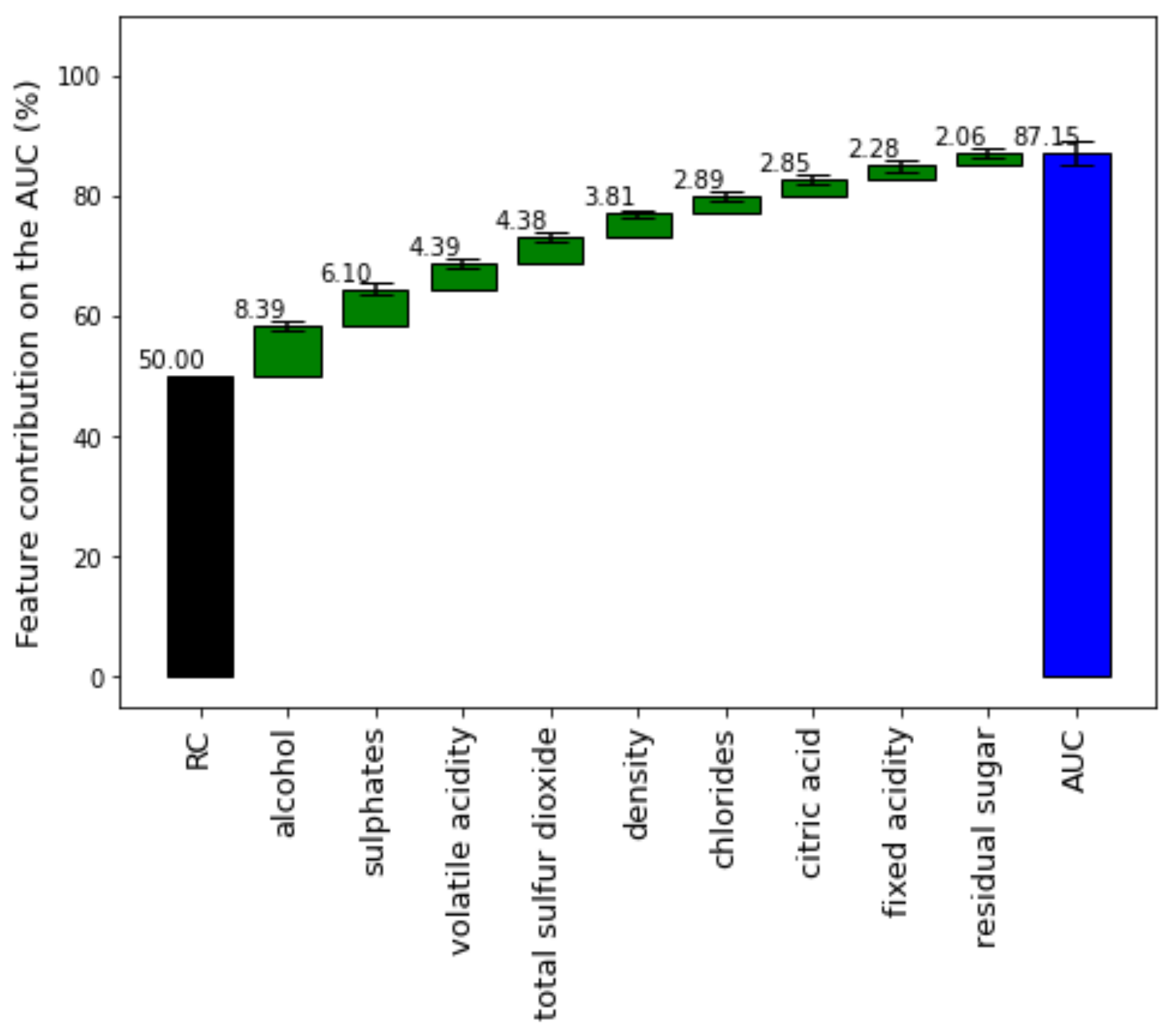}
\label{fig:robustness_auc_int_featSelec_wine}}
\hfill
\caption{Application of ShapAUC as a feature selection method - Red Wine Quality dataset.}
\label{fig:robustness_featSelec_wine}
\end{figure}

\begin{figure}[!ht]
\centering
\subfloat[With all features.]{\includegraphics[width=3.0in]{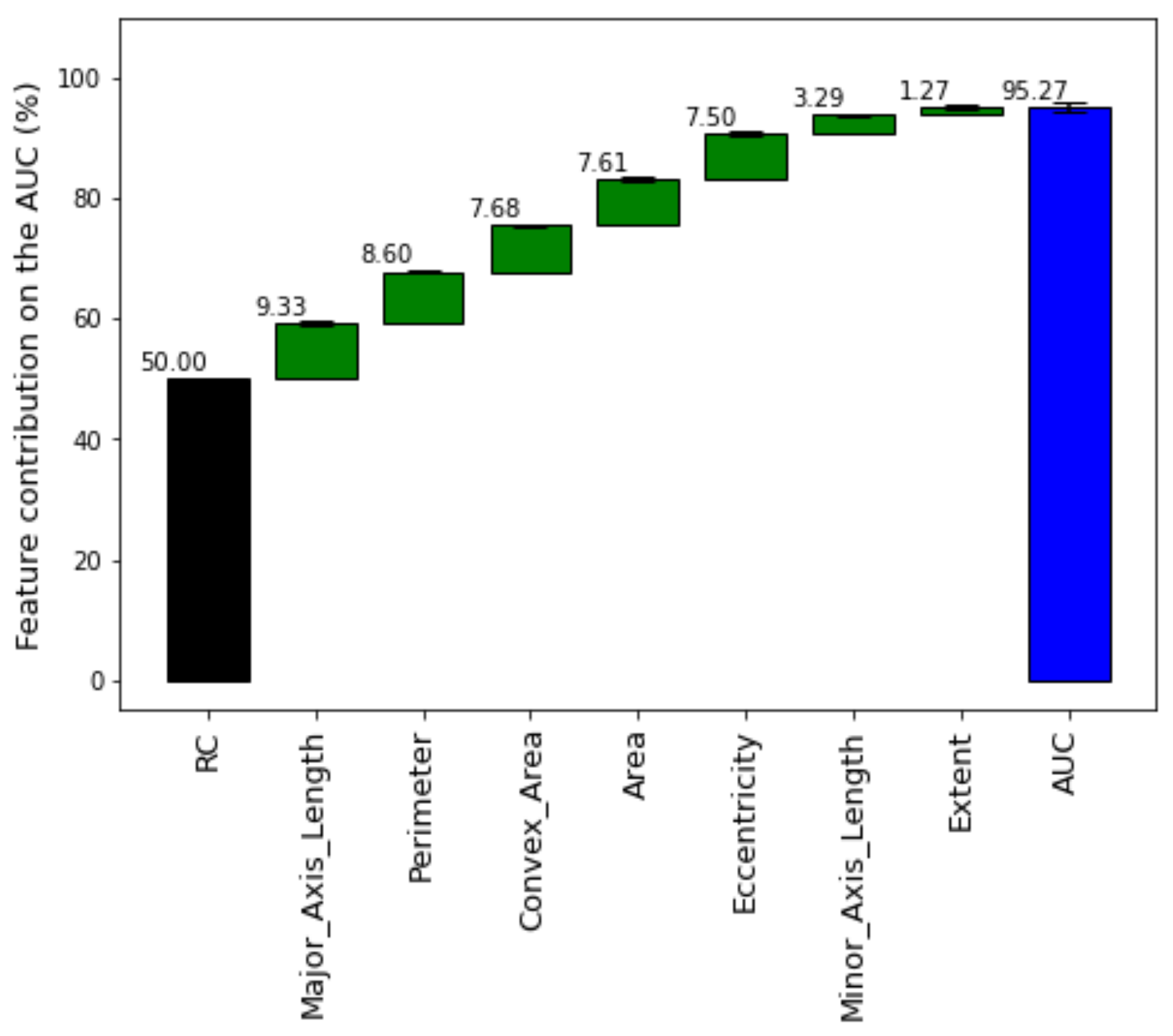}
\label{fig:robustness_auc_int_rice}}
\hfill
\subfloat[Removing \textit{Extent}.]{\includegraphics[width=3.0in]{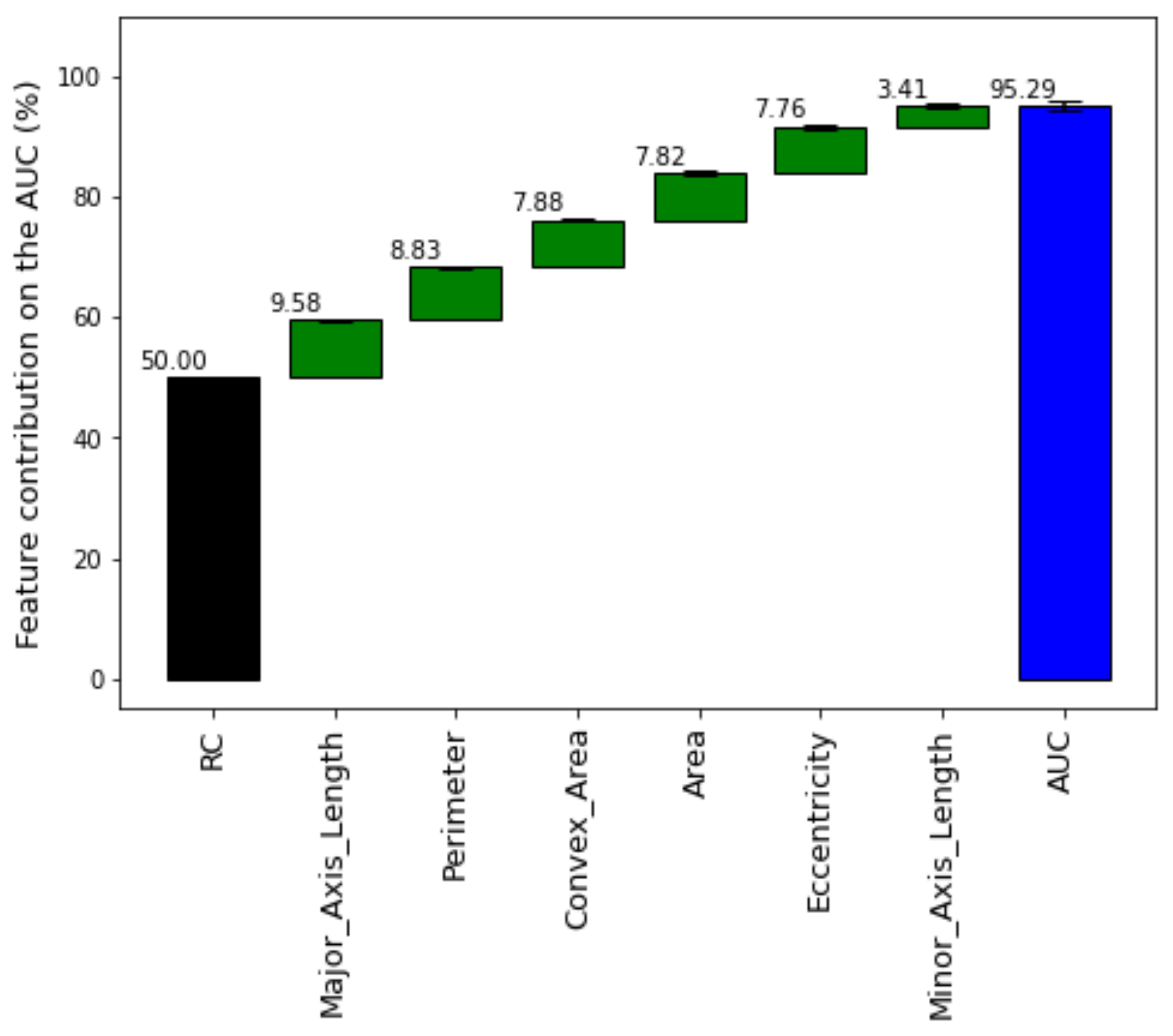}
\label{fig:robustness_auc_int_featSelec_rice}}
\hfill
\caption{Application of ShapAUC as a feature selection method - Rice dataset.}
\label{fig:robustness_featSelec_rice}
\end{figure}

\begin{figure}[!ht]
\centering
\subfloat[With all features.]{\includegraphics[width=3.0in]{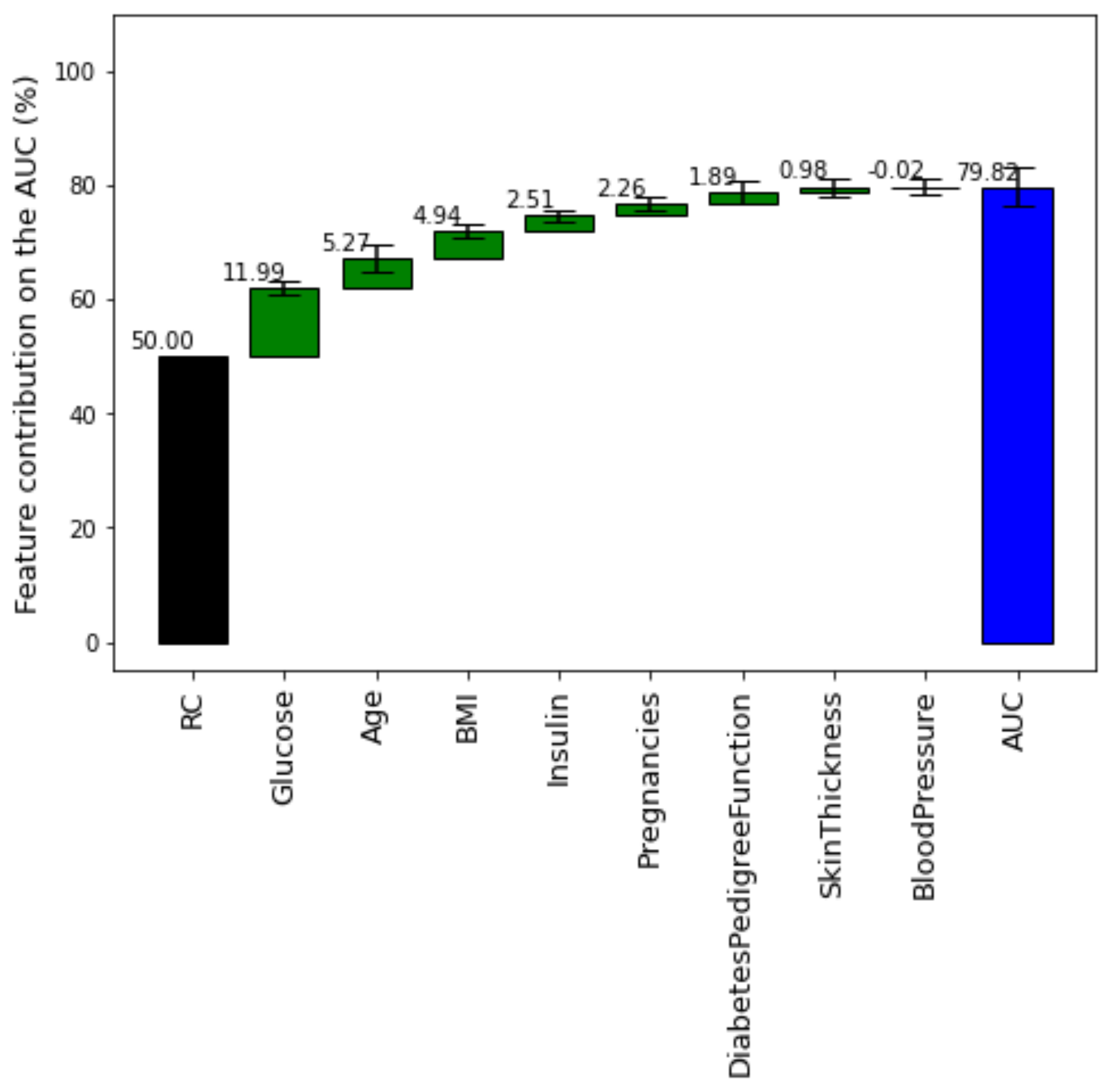}
\label{fig:robustness_auc_int_diabetes}}
\hfill
\subfloat[Removing \textit{Glucose}.]{\includegraphics[width=3.0in]{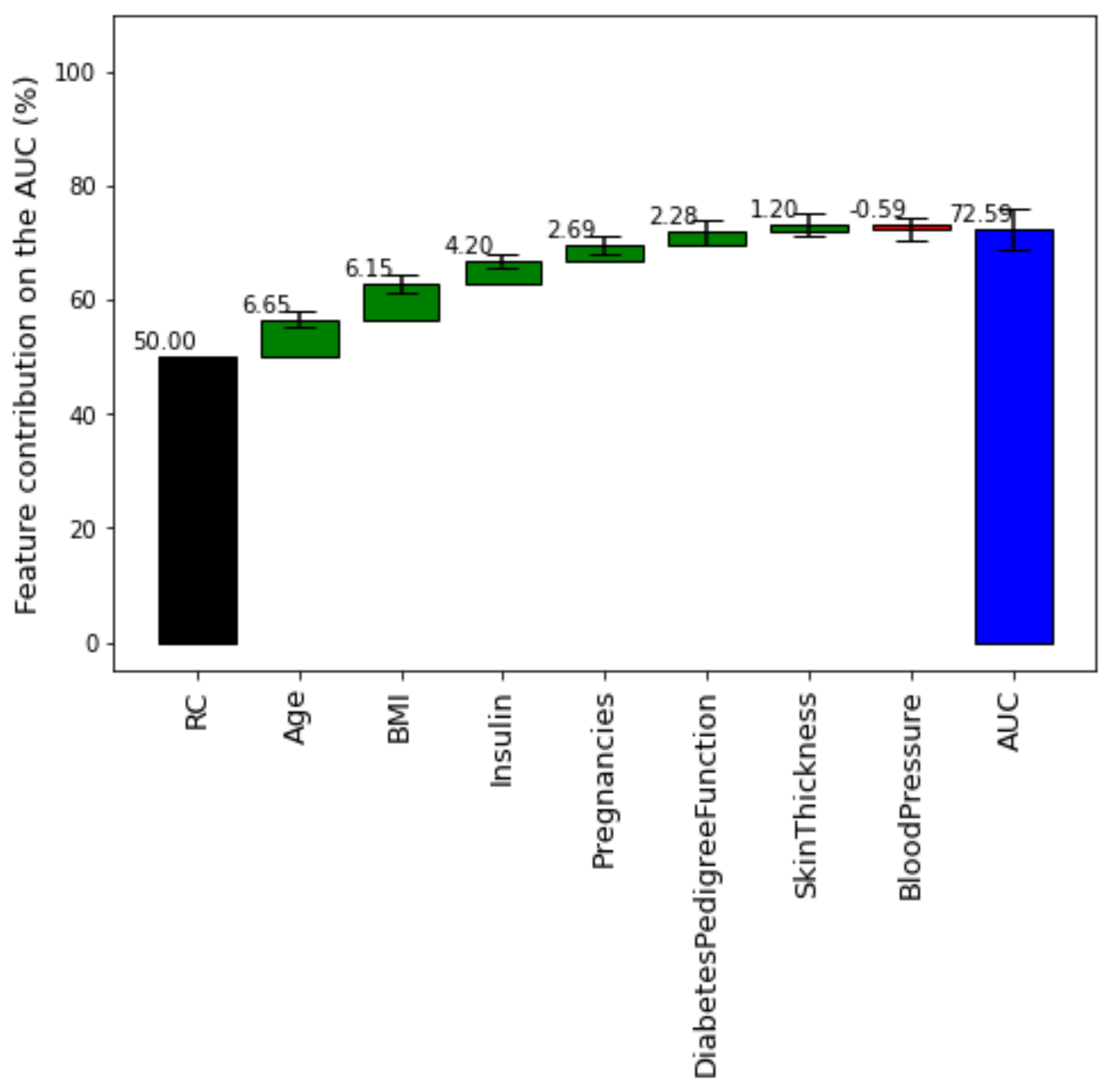}
\label{fig:robustness_auc_int_featSelec_diabetes}}
\hfill
\caption{Application of ShapAUC as a feature selection method - Pima Indians Diabetes dataset.}
\label{fig:robustness_featSelec_diabetes}
\end{figure}

\end{document}